\documentclass[10pt,twocolumn,letterpaper]{article}
\usepackage[T1]{fontenc}   %
\usepackage[pagenumbers]{cvpr}      %

\renewcommand{\paragraph}[1]{\par\vspace{.5em}\noindent\textbf{#1}}

\usepackage{colortbl}
\definecolor{oursbg}{RGB}{255,232,204}
\AtBeginDocument{\crefname{appendix}{Appendix}{Appendices}\Crefname{appendix}{Appendix}{Appendices}\crefname{subappendix}{Appendix}{Appendices}\Crefname{subappendix}{Appendix}{Appendices}}

\usepackage{amsmath,amssymb,booktabs,multirow,graphicx,xspace}
\usepackage{makecell}
\usepackage{placeins}   %

\definecolor{cvprblue}{rgb}{0.21,0.49,0.74}
\usepackage[pagebackref,breaklinks,colorlinks,allcolors=cvprblue]{hyperref}

\title{Static Content First, Dynamics on Top:\\ Refining the Mean for Text-Video Retrieval}  %

\author{Toya Oyama$^{1,2}$ \qquad Rainer Lienhart$^{3}$ \qquad Shin'ichi Satoh$^{2,1}$\\
$^{1}$The University of Tokyo \qquad $^{2}$National Institute of Informatics \qquad $^{3}$University of Augsburg}

\begin{document}
\twocolumn[{\renewcommand\twocolumn[1][]{#1}\maketitle\vspace{-2em}\providecommand{\capicon}[1]{\protect\raisebox{-0.12em}{\protect\includegraphics[height=0.85em]{figs/#1.pdf}}}
\begin{center}
  \begin{minipage}[b]{0.70\textwidth}\centering
    \includegraphics[height=1.40in]{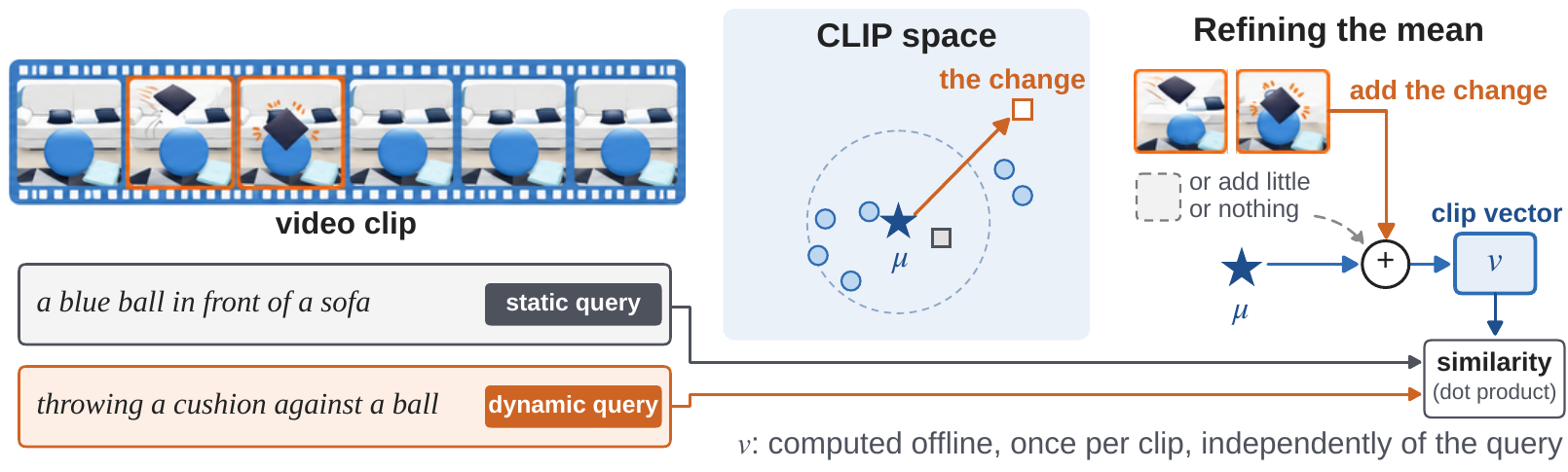}\\[1pt]
    {\small (a) Concept (schematic)}
  \end{minipage}\hfill
  \begin{minipage}[b]{0.29\textwidth}\centering
    \includegraphics[height=1.40in]{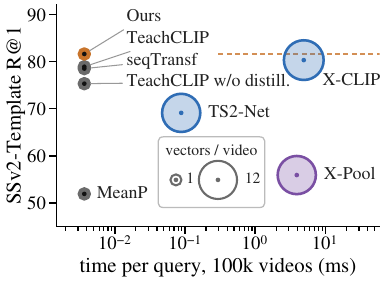}\\[1pt]  %
    {\small (b) Cost and accuracy}
  \end{minipage}
  \captionof{figure}{\textbf{Static content first, dynamics on top.}
  (a)~In the CLIP space, the frame embeddings (\capicon{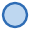}) of a video clip gather around their mean $\mu$ (\capicon{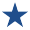}), and what changes over time is a small deviation from it.
  For a static query (\capicon{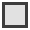}), which asks about what all frames show, the clip is retrieved even when its clip vector $v$ equals $\mu$. A dynamic query (\capicon{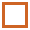}), which asks about what changes over time, also needs the change to be encoded in the clip vector.
  We add the change to $\mu$ offline, once for each clip in the database and independently of any text query; each text query is matched to the resulting single vector by one dot product.
  (b)~Measured retrieval time per query over $100$k videos (A40, fp16, log scale) vs.\ R@1 on SSv2-Template; marker area: vectors stored per video. Single-vector methods (left) are about $1{,}000\times$ faster than X-CLIP and X-Pool.
  }
  \label{fig:teaser}
\end{center}
\vspace{0.6em}
}]   %
\begin{abstract}
The fastest text-video retrieval models built on CLIP turn a video clip into a single vector, which is computed offline and compared with the query vector by one dot product.
The frames of a clip are highly similar, so their mean holds most of the content; changes over time are small deviations from it.
Prior methods such as CLIP4Clip feed frame embeddings directly to a temporal transformer and may miss how the clip evolves, because they are dominated by their shared content.
Instead, we separate the frame embeddings into their mean and the deviations, and feed only the deviations to a temporal transformer.
Starting from this mean, the clip vector attends to the transformer outputs to add what it still lacks, and a null slot lets it add little or nothing when the clip is mostly static.
This runs once per clip, independently of the query, so each clip stays one vector that holds both static and dynamic content.
On Something-Something v2, whose queries describe actions, our method outperforms every baseline. Taking the better of its results with and without distillation, it is the best on six of ten benchmarks and within $0.8$ R@1 of the best on the others. It retrieves about $1{,}000\times$ faster than the heaviest models.
\end{abstract}

\section{Introduction}
\label{sec:intro}

Text-video retrieval ranks videos based on a text query.
Most recent methods build on CLIP~\cite{radford2021clip} and differ in how they turn the per-frame embeddings into a representation of the video clip.
The fastest family encodes each video clip into one vector, computed offline, and compares it with the text vector by a single dot product;
we call such models \emph{single-vector}.  %
Mean pooling of the frame embeddings is already a strong single-vector baseline.
The standard way to add time is to run a temporal transformer over the frame embeddings before pooling them (seqTransf)~\cite{luo2022clip4clip}. Its CLIP4Clip form, which averages the outputs, is CLIP4Clip-seqTransf (\cref{sec:seqtransf}).
TeachCLIP~\cite{tian2024teachclip} even argues that one vector summarizing the whole clip is almost sufficient.  %

This strength comes from what a video clip is.
A clip usually stays in one scene. Its frames are highly similar in meaning and appearance, and only the motion changes, little by little.
In the CLIP space the frame embeddings therefore gather tightly around their mean (\cref{fig:teaser}a), and the change is a small deviation from it.
Some queries ask exactly about this change, such as \emph{throwing something against something} on Something-Something~v2~\cite{goyal2017ssv2,lei2023singularity}.
A good clip representation thus needs both the \emph{static} scene content that all frames share and the \emph{dynamic} content, the temporal change on top of it.
We call queries about the shared content of contiguous frames static queries and queries about their temporal change dynamic queries.

Existing single-vector methods keep the shared content and capture the change only in part.
Mean pooling keeps only the shared content.
seqTransf adds to each frame a correction that carries information about the whole clip but has almost the same direction for every frame. Each attention layer passes the mean of the frame embeddings to every frame with full weight,
and the deviations that carry the change are only a small part of each frame (\cref{sec:analysis}).  %
TeachCLIP pools these outputs with weights that a small network predicts from each frame, but it chooses among frames whose corrections point the same way.
Late-interaction methods~\cite{ma2022xclip,reddy2025videocolbert} keep several vectors per clip, such as one per frame, and compare each of them with the text and its words at query time. Text-conditioned methods~\cite{gorti2022xpool} let the text query decide how the frame embeddings are pooled.
Both make retrieval about $1{,}000\times$ slower: over $100$k videos, a query takes $4$\,ms instead of $0.004$\,ms (\cref{fig:teaser}b).  %

We keep one vector per clip and refine it additively, once per clip and independently of any text query.
Our method keeps the mean vector $\mu$ of the normalized frame embeddings unchanged as the base.
A temporal transformer, the same as in seqTransf, reads the deviations of the frame embeddings from their plain mean, so it sees what changes rather than what stays, and turns them into one refinement candidate per frame.
The mean vector queries these candidates with a cross-attention (\emph{mean-queried attention}) and adds what it still lacks; repeating this step refines the vector further.
A zero-valued extra candidate, the \emph{null slot}, lets it add little or nothing when the mean vector suffices.  %

We retrained every method under the same setting on ten benchmarks. Our method outperforms every baseline on Something-Something v2, whose queries describe actions. This includes the late-interaction X-CLIP, which stores $13$ vectors per clip: $81.6$ vs.\ $80.3$ R@1 on SSv2-Template and $51.5$ vs.\ $48.7$ on SSv2-Label.
Across all ten benchmarks, taking the better of its results with and without distillation, it is the best on six and within $0.8$ R@1 of the best on the other four.
With the distillation loss of TeachCLIP, it also outperforms every baseline on MSR-VTT, including the heavier X-CLIP and X-Pool, and on SSv2-Template it exceeds its X-CLIP teacher, which TeachCLIP does not.
It also transfers better to benchmarks it was not trained on: trained on MSR-VTT and tested on five others, or trained on SSv2-Template and tested on MSR-VTT,
it outperforms every baseline other than X-Pool (\cref{tab:xdomain}).
Our contributions are:
\begin{itemize}
  \item An analysis of single-vector video-clip representations exploiting CLIP features: the frame embeddings of a clip are highly similar to each other,
        so a temporal transformer on frame embeddings adds corrections aligned across frames. When the query asks about the change, reweighting or selecting frames helps little (\cref{sec:analysis}).
  \item Mean-queried attention with a null slot refines the mean vector additively and adds little or nothing when $\mu$ suffices, at the retrieval cost of mean pooling (\cref{sec:method}).
  \item Experiments on ten benchmarks under one setting: the best R@1 on Something-Something v2 and the best or within $0.8$ R@1 of the best on the others,
        and, with distillation, one vector per clip that exceeds its X-CLIP teacher on SSv2-Template and every baseline on MSR-VTT (\cref{sec:experiments}).
\end{itemize}

\section{Related Work}
\label{sec:related}

\paragraph{CLIP-based text-video retrieval.}
Most recent methods building on CLIP~\cite{radford2021clip} differ in how they aggregate the frame embeddings.
CLIP4Clip~\cite{luo2022clip4clip} found mean pooling a strong baseline against temporal heads. CLIP2Video~\cite{fang2021clip2video} adds difference tokens, and TS2-Net~\cite{liu2022ts2net} shifts tokens inside the image encoder and weights its frames by the text at test time.
TeachCLIP~\cite{tian2024teachclip} adds two parts to seqTransf: an attentional frame-feature aggregation (AFA) block, which pools the transformer outputs with weights predicted from each frame by an MLP,
and distillation from heavy teachers, which also supervises these weights.
CLIP-ViP~\cite{xue2023clipvip} instead adapts the image encoder by video-language post-pretraining, i.e., a further pretraining stage on video-text data.
We stay single-vector like TeachCLIP but, instead of pooling the transformer outputs by weights, keep the mean vector and add weighted refinement candidates.

\paragraph{Temporal modeling on image features.}
Light temporal modules on image backbones add their output to the CLIP features (CLIP4Clip, STAN~\cite{liu2023stan}), read them with learnable queries (EVL~\cite{lin2022evl}), or make motion explicit by differences between neighboring frames (CLIP2Video, TDN~\cite{wang2021tdn}).
We instead subtract the plain mean of the clip and keep the mean vector as the base and the attention query.  %
Video-ColBERT also separates a static from a temporal stream, but scores them by matching many vectors per clip.
The attention pooling of CLIP's ResNet encoder also queries with the mean~\cite{radford2021clip}, but its output replaces the mean, with neither a residual nor a null slot.

\paragraph{Dual encoders, late interaction and cross encoders.}
A \emph{dual encoder} maps the video and the text to one vector each and scores a pair by one dot product,
so the clip vectors can be computed offline and indexed; single-vector models such as mean pooling, seqTransf and TeachCLIP are dual encoders.
\emph{Late interaction}~\cite{khattab2020colbert} keeps several vectors per side and combines many token-level similarities, as in X-CLIP~\cite{ma2022xclip} and Video-ColBERT~\cite{reddy2025videocolbert}.
A \emph{cross encoder} lets the text enter the video encoding; text-conditioned pooling such as X-Pool~\cite{gorti2022xpool} is a light form of it.
Our method is a dual encoder: its cross-attention runs in the video encoder, between the mean vector and the refinement candidates of the same clip, so a pair is still scored by one dot product.

\paragraph{Attention sinks and null options.}
The weights of softmax attention always sum to one, so an attention head cannot add nothing unless it can put its weight somewhere harmless.
Models learn such places themselves: attention sinks in language models~\cite{xiao2024streamingllm}, outlier activations of heads that try to do nothing~\cite{bondarenko2023quantizable}, and high-norm background patches in vision transformers, for which register tokens add dedicated slots~\cite{darcet2024registers}.
A slot with a zero value gives this place explicitly, as in the off-by-one softmax~\cite{miller2023offbyone} and the per-head sink logits of gpt-oss~\cite{openai2025gptoss}.
Our null slot is such a slot, but its logit depends on the attention query, so how much is added to the mean vector is decided for each clip.

\paragraph{Temporally heavy benchmarks.}
On many video-language benchmarks, models that see single frames reach strong results~\cite{buch2022atp,lei2023singularity}, because their queries mostly describe what is visible rather than how it changes.
Lei~\etal~\cite{lei2023singularity} therefore built retrieval on Something-Something v2 (SSv2)~\cite{goyal2017ssv2}, where appearance alone does not decide the answer. Test of Time~\cite{bagad2023testoftime} and RTime~\cite{du2024rtime} probe before/after relations and time reversal.
We evaluate on SSv2, RTime, Charades~\cite{sigurdsson2016charades} and standard benchmarks.

\section{What the Mean Vector of Frame Embeddings Misses}
\label{sec:analysis}

We first examine how CLIP-based retrieval models summarize a video clip.
We use CLIP ViT-B/32~\cite{radford2021clip,dosovitskiy2021vit} and decode each clip at $1$ frame per second ($6$ on SSv2, whose clips last a few seconds); as in CLIP4Clip~\cite{luo2022clip4clip}, we sample up to $T=12$ of these frames uniformly over the clip, and shorter clips use all their frames. We study MSR-VTT~\cite{xu2016msrvtt}, where most queries describe appearance~\cite{lei2023singularity,buch2022atp}, and SSv2 retrieval~\cite{lei2023singularity} (SSv2-Template and SSv2-Label, SSv2-T/-L; \cref{sec:experiments}), where queries describe actions such as \emph{pushing something from left to right}.
Each subsection ends with a design principle that \cref{sec:method} follows.

\paragraph{Notation.}
\label{sec:prelim}
CLIP encodes the $T$ frames of a clip independently into embeddings $x_1,\dots,x_T\in\mathbb{R}^d$; we write $x=(x_1,\dots,x_T)$ for the sequence and $\bar{x}=\frac{1}{T}\sum_t x_t$ for its plain mean.
The text query is encoded by the CLIP text encoder into a single vector $s\in\mathbb{R}^d$, and videos are ranked by the cosine similarity between $s$ and a single clip vector $v$, which each method derives from $x$.
Mean pooling (MeanP; CLIP4Clip-MeanP in the tables) uses the average of the normalized frame embeddings, $\mu=\frac{1}{T}\sum_t x_t/\lVert x_t\rVert$.  %

\subsection{Frames are highly similar}
\label{sec:frames}
We measure the mean cosine similarity between the 12 frame embeddings of a clip, and between the word features of its query from the CLIP text encoder,
after subtracting the dataset mean of each modality.
The centering removes the anisotropy of the CLIP space, so unrelated items score about zero.
Frame embeddings of the same clip remain highly similar ($0.64$ on MSR-VTT, $0.71$ on SSv2), while words of the same sentence are not ($0.18$ and $0.20$; example in \cref{sec:supp_similarity}).
Frame embeddings are thus about $3.5\times$ more similar to each other than words are ($2.6$--$4.6\times$ on seven other benchmarks).
The mean vector already describes most of the content, and what changes is only a small part of each frame embedding.

\paragraph{Principle 1.} \emph{Keep the mean vector of the frame embeddings as the base of the clip representation.}

\subsection{A temporal transformer adds one shared direction to the clip}
\label{sec:seqtransf}
CLIP4Clip applies a transformer~\cite{vaswani2017attention} over the frame embeddings $x_t$, with position embeddings $p=(p_1,\dots,p_T)$ and a residual connection (seqTransf),\footnote{The residual connection and the initialization of $\mathrm{Transformer}$ from the first layers of the CLIP text encoder follow the official implementation of CLIP4Clip. $\mathrm{Transformer}(\cdot)$ includes the residual connections of its own layers, so it returns its input plus what the layers add.}
and averages its outputs over time (CLIP4Clip-seqTransf).
The AFA block of TeachCLIP~\cite{tian2024teachclip} instead weights them with $w_t=\mathrm{softmax}_t(\mathrm{MLP}(h_t)/\tau)$, $\tau=0.1$, where $h_t=x_t+\mathrm{Transformer}(x+p)_t$ is the output for frame $t$. seqTransf with AFA, trained without distillation, is the TeachCLIP student (TeachCLIP w/o distillation). We write seqTransf when a statement holds for both pooling choices.
Before the per-frame normalization, the pooled output is
\begin{equation}
  \begin{split}
    &\sum_{t=1}^{T}w_t\bigl(x_t+\mathrm{Transformer}(x+p)_t\bigr)\\
    &\quad=\tilde{x}+\sum_{t=1}^{T}w_t\,\mathrm{Transformer}(x+p)_t,
  \end{split}
  \label{eq:seqtransf}
\end{equation}
where $\tilde{x}=\sum_t w_t x_t$ is the weighted mean of the frame embeddings ($\bar{x}$ for uniform weights).
Let $\Delta_t$ be what the layers add to frame $t$, so $\mathrm{Transformer}(x+p)_t=x_t+p_t+\Delta_t$. Ignoring positions, the pooled output is then twice a reweighted mean of the frame embeddings plus $\sum_t w_t\Delta_t$.
This correction carries information about the whole clip: it lifts SSv2-Template R@1 from $51.9$ for MeanP to $78.5$ for CLIP4Clip-seqTransf, although only from $43.4$ to $44.2$ on MSR-VTT.
But the corrections of the frames point in almost the same direction, so they largely reinforce one another when averaged.
We measure this with
\begin{equation}
  \rho=\frac{\bigl\lVert\frac{1}{T}\sum_{t}\Delta_t\bigr\rVert}{\frac{1}{T}\sum_{t}\lVert\Delta_t\rVert}\in[0,1],
  \label{eq:rho}
\end{equation}
which is $1$ when all $\Delta_t$ point in the same direction. We find $\rho=0.98$ on MSR-VTT, $0.96$ on SSv2-Label and $0.94$ on SSv2-Template (TeachCLIP w/o distillation).
The reason is that the attention weights of a frame sum to one, so each layer passes the mean of its inputs to every frame in full. The weights act only on the deviations $x_t-\bar{x}$, which hold about $30\%$ of the squared norm of a frame embedding after removing the dataset mean.

Reweighting these frames helps a little when queries describe appearance, but not where the change matters. On SSv2-Template, the AFA weights focus on about $3$ of $12$ frames, yet TeachCLIP w/o distillation ($75.3$) stays below CLIP4Clip-seqTransf ($78.5$), which averages them uniformly.
The input helps more: feeding only the deviations lifts TeachCLIP w/o distillation from $75.3$ to $82.2$ on SSv2-Template (\cref{sec:ablation}).

\paragraph{Principle 2.} \emph{Feed the temporal module the deviations from the plain mean, so that it can use what differs between frame embeddings.}

\subsection{Selection does not recover what changes}
\label{sec:selection}
Selecting the frames that matter works when the query asks about what is visible. X-Pool~\cite{gorti2022xpool}, which weights the frames by their relevance to the query, is above MeanP on all six benchmarks whose queries mostly describe appearance (\cref{tab:standard}).
Selection does not work when the query asks about a change. On SSv2-Template X-Pool reaches only $55.9$ R@1, far below CLIP4Clip-seqTransf ($78.5$), and it is below MeanP on SSv2-Label and Charades (\cref{tab:specific}).
A change is not contained in any single frame.
What we add to the mean vector must also be optional. When the mean vector already describes the clip, adding the change can hurt: training without the null slot lowers R@1 on both MSR-VTT and SSv2-T (\cref{tab:abl_parts}), and forcing a trained model to always add the full amount costs far more (\cref{sec:why}).

\paragraph{Principle 3.} \emph{Instead of selecting frames, let the mean vector ask for what it lacks and add only that, with the option to add nothing.}

\begin{figure*}[t]
  \centering
  \includegraphics[width=\linewidth]{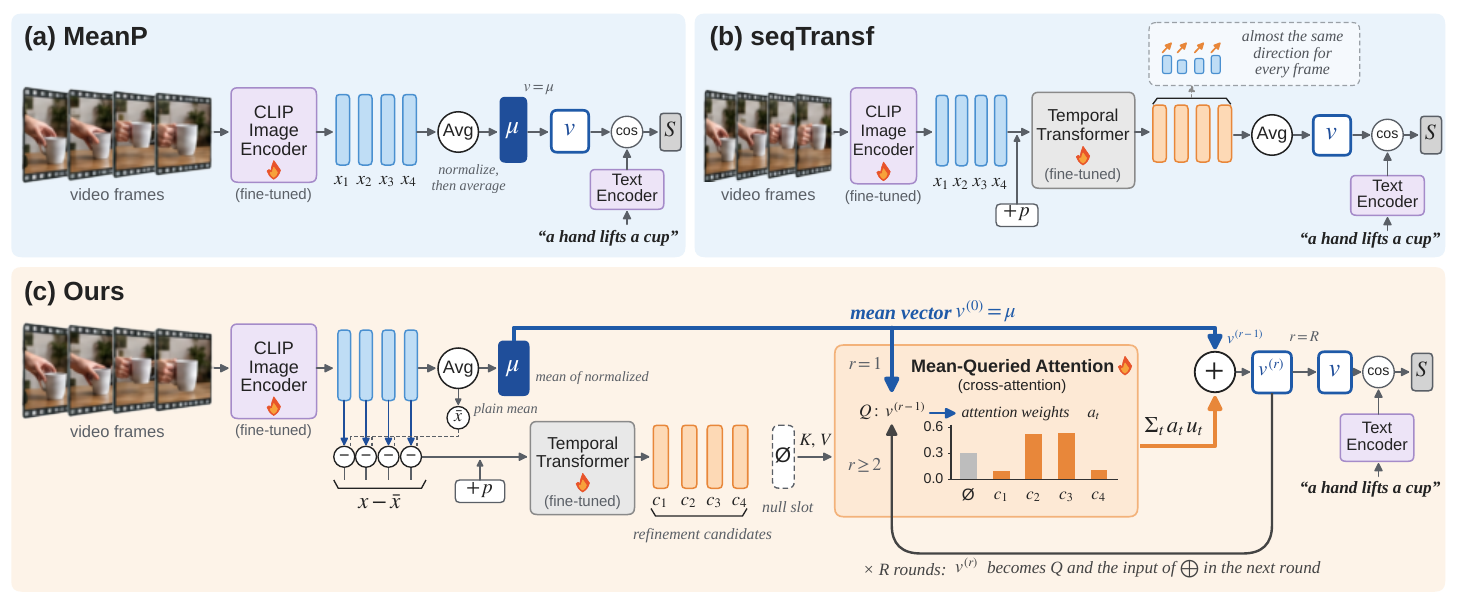}   %
  \caption{\textbf{From mean pooling to our method.}
  (a)~MeanP averages the normalized frame embeddings into $\mu$.
  (b)~seqTransf (CLIP4Clip) runs a temporal transformer over them and averages (or, in TeachCLIP, weights) its outputs; the vectors it adds share almost one direction, acting as one correction.
  (c)~Ours keeps $\mu$ as the base. A temporal transformer reads the deviations $x-\bar{x}$ from the plain mean $\bar{x}$ (not $\mu$) and suggests the dynamic changes as refinement candidates $c_t$.
  Mean-queried attention adds a weighted sum: its query is the current vector ($v^{(0)}=\mu$ in round 1); keys $k_t$ and values $u_t$ come from the candidates and a null slot (value $0$), which lets it add nothing. Each round adds to the previous vector $v^{(r-1)}$, giving $v^{(r)}$; finally $v=v^{(R)}$.
  The cos node gives the score $S$, the cosine of $v$ and the text embedding.}
  \label{fig:method}
\end{figure*}

\section{Method}
\label{sec:method}

Our method follows the three principles of \cref{sec:analysis}.
It keeps the mean vector $\mu$ of the normalized frame embeddings unchanged as the base, so everything it adds is a residual that keeps the strong signal of mean pooling (Principle~1),
builds refinement candidates from the deviations of each frame embedding from the plain mean (Principle~2),
and lets the mean vector query these candidates and add only what it lacks, with the option to add nothing (Principle~3).
\cref{fig:method} compares it with mean pooling and seqTransf~\cite{luo2022clip4clip}.

\paragraph{Refinement candidates.}
To let the temporal transformer focus on what changes, we remove the plain mean before it:
\begin{equation}
  c_t=\mathrm{Transformer}\bigl(x-\bar{x}+p\bigr)_t,\qquad t=1,\dots,T.
  \label{eq:cand}
\end{equation}
Since all frame embeddings share most of their content, $x_t-\bar{x}$ keeps only the frame-specific part, and the plain mean is no longer passed on to every frame.
The transformer has the same architecture and initialization as in seqTransf, so \cref{eq:cand} adds no parameters.
We call $c_1,\dots,c_T$ the \emph{refinement candidates} (candidates for short).

\paragraph{Mean-queried attention with a null slot.}
\label{sec:query}
The mean vector decides which candidates to add.
We use $\mu$ as the single query of a cross-attention over the candidates, which we call \emph{mean-queried attention} (after the first round, the query is the refined vector; \cref{eq:rounds}):
\begin{equation}
  \begin{gathered}
    q=W_q\,\mathrm{LN}(\mu),\qquad k_t=W_k\,\mathrm{LN}(c_t),\\
    u_t=\frac{\lVert\mu\rVert}{\sqrt{d}}\,\mathrm{LN}(W_v c_t),
  \end{gathered}
  \label{eq:qkv}
\end{equation}
where $\mathrm{LN}$ is layer normalization and $W_q,W_k,W_v\in\mathbb{R}^{d\times d}$ are learned projections.
The value $u_t$ has no learned scale, so its norm is about $\lVert\mu\rVert$.
We add to the $T$ candidates a \emph{null slot} with index $0$, an extra candidate whose value is zero, so attending to it adds nothing.
Its logit is $l_0=q^\top k_0/\sqrt{d}+b$ with a learned key $k_0$ and bias $b$, while the candidates have $l_t=q^\top k_t/\sqrt{d}$.
The weights are a softmax over the candidates and the null slot,
\begin{equation}
  \begin{gathered}
    (a_0,a_1,\dots,a_T)=\mathrm{softmax}(l_0,l_1,\dots,l_T),\\
    v^{(1)}=\mu+\sum_{t=1}^{T}a_t\,u_t .
  \end{gathered}
  \label{eq:add}
\end{equation}
The amount added to $\mu$ is the total candidate weight,
\begin{equation}
  \sum_{t=1}^{T}a_t=1-a_0=\sigma\Bigl(\log\sum_{t=1}^{T}e^{l_t}-l_0\Bigr),
  \label{eq:amount}
\end{equation}
where $\sigma$ is the sigmoid.
Because every value has about the norm of $\mu$, how much is added is set by this amount, through the competition with the null slot, and not by the scale of the values.
This amount decides, for each clip, whether the mean vector already describes the clip well, so that little or nothing needs to be added, or whether an essential part of the clip's meaning lies in a change that needs to be added.
We initialize $W_v$ to the identity and set $b$ so that about $10\%$ is added at the start. Training thus begins close to MeanP.
Unlike EVL~\cite{lin2022evl}, whose learned query starts the same for every video, our query starts from the clip's own mean.

\paragraph{Iterative refinement.}
\label{sec:rounds}
One application of \cref{eq:add}, which we call a \emph{round}, adds a weighted sum of the values, with weights that sum to $1-a_0$.
We repeat it $R$ times, and each round adds to the vector left by the previous one.
With several rounds, later rounds add more where dynamic content matters, mostly from the same frames; training with one, two or four rounds still gives similar R@1 on SSv2-T (\cref{sec:ablation}).
Round $r$ has its own parameters and is queried by the current vector, $q^{(r)}=W_q^{(r)}\mathrm{LN}(v^{(r-1)})$ with $v^{(0)}=\mu$, so its query reflects what earlier rounds added:
\begin{equation}
  v^{(r)}=v^{(r-1)}+\sum_{t=1}^{T}a^{(r)}_t\,u^{(r)}_t .
  \label{eq:rounds}
\end{equation}
Here $a^{(r)}_t$ and $u^{(r)}_t$ are computed as in \cref{eq:qkv,eq:add} with the parameters of round $r$, and all rounds share the same candidates $c_t$.
We fix $R=2$ on every benchmark, although other values can be better on some benchmarks; longer videos with several actions gain from more rounds (\cref{sec:supp_rounds}).  %

\paragraph{Training and complexity.}
The final clip vector $v^{(R)}$ is $\ell_2$-normalized and trained with the standard symmetric InfoNCE loss~\cite{oord2018cpc} of CLIP4Clip. We add no other loss.
Each round adds about $0.8$M parameters (three projections, two layer normalizations, the null key and bias).

\section{Experiments}
\label{sec:experiments}
\begin{table*}[t]
  \caption{Text-to-video retrieval on benchmarks that need dynamic content. All methods use CLIP ViT-B/32 and are trained by us on the
  same frames; mean over three seeds; the standard deviation of R@1 is at most $1.3$, and up to $4.3$ on SSv2-Template, where one query is $0.57$ R@1 (per seed: \cref{sec:supp_seeds}). TeachCLIP and Ours + distillation use the same teachers (Sec.~\ref{sec:experiments}). SSv2-Template / Label follow the Singularity
  protocol~\cite{lei2023singularity}. RTime (Hard): the forward and the reversed captions are both queries, and the gallery holds every clip and its reversed copy. Ours: $R{=}2$ rounds (\cref{sec:rounds}) on every benchmark (more rounds: \cref{sec:supp_rounds}). Bold: best per column.}
  \centering
  \small
  \setlength{\tabcolsep}{4pt}
  \renewcommand{\arraystretch}{0.92}  %
  \begin{tabular}{lcccccccccccc}
    \toprule
    Method & \multicolumn{3}{c}{SSv2-Template} & \multicolumn{3}{c}{SSv2-Label} & \multicolumn{3}{c}{Charades} & \multicolumn{3}{c}{RTime (Hard)} \\
    \cmidrule(lr){2-4} \cmidrule(lr){5-7} \cmidrule(lr){8-10} \cmidrule(lr){11-13}
    & R@1 & R@5 & R@10 & R@1 & R@5 & R@10 & R@1 & R@5 & R@10 & R@1 & R@5 & R@10 \\
    \midrule
    CLIP4Clip-MeanP & 51.9 & 87.6 & 96.2 & 39.7 & 67.1 & 76.5 & 15.0 & 32.3 & 41.7 & 34.8 & 84.4 & 91.4 \\
    CLIP4Clip-seqTransf & 78.5 & 97.5 & 99.4 & 48.2 & 74.8 & 82.5 & \textbf{16.9} & \textbf{36.2} & \textbf{46.9} & 35.1 & 83.9 & 91.8 \\
    TeachCLIP w/o distill. & 75.3 & 97.9 & 99.2 & 48.0 & 75.2 & 83.4 & 16.6 & 35.6 & 46.2 & 35.5 & 84.5 & 91.8 \\
    X-CLIP & 80.3 & \textbf{98.3} & 99.4 & 48.7 & 74.8 & 83.1 & 15.7 & 34.3 & 44.2 & 34.1 & 81.6 & 89.7 \\
    X-Pool & 55.9 & 91.6 & 97.1 & 38.2 & 66.9 & 77.1 & 12.1 & 29.0 & 39.4 & 35.2 & \textbf{85.7} & 92.4 \\
    TS2-Net & 69.1 & 95.2 & 98.5 & 46.3 & 74.6 & 82.3 & 14.3 & 32.8 & 42.5 & 34.4 & 83.7 & 91.3 \\
    TeachCLIP & 78.9 & \textbf{98.3} & 99.2 & 48.4 & 75.2 & 83.4 & 15.5 & 34.1 & 43.8 & 32.5 & 82.0 & 89.9 \\
    \midrule
    \rowcolor{oursbg} Ours & 81.6 & 97.9 & \textbf{99.6} & \textbf{51.5} & 76.3 & 84.0 & 16.6 & 35.8 & 46.4 & 35.5 & 84.7 & 91.8 \\
    \rowcolor{oursbg} Ours + distillation & \textbf{82.2} & 97.9 & 99.4 & 50.7 & \textbf{77.1} & \textbf{84.9} & 16.3 & 34.9 & 45.3 & \textbf{36.3} & 85.6 & \textbf{92.5} \\
    \bottomrule
  \end{tabular}
  \label{tab:specific}
\end{table*}

\begin{table*}[t]
  \caption{Text-to-video retrieval on standard benchmarks. Setting as in \cref{tab:specific}. TeachCLIP on MSR-VTT is the original implementation; elsewhere, its student and losses in our code. Ours + distillation uses only the coarse-grained distillation loss of TeachCLIP.
  $^\ast$VATEX: the 1,303 test videos of the HGR split~\cite{chen2020hgr} that are still available, the same for all methods. Bold: best per column.}
  \centering
  \footnotesize
  \setlength{\tabcolsep}{2.2pt}
  \renewcommand{\arraystretch}{0.92}  %
  \begin{tabular}{lcccccccccccccccccc}
    \toprule
    Method & \multicolumn{3}{c}{MSR-VTT} & \multicolumn{3}{c}{ActivityNet} & \multicolumn{3}{c}{MSVD} & \multicolumn{3}{c}{VATEX$^\ast$} & \multicolumn{3}{c}{LSMDC} & \multicolumn{3}{c}{DiDeMo} \\
    \cmidrule(lr){2-4} \cmidrule(lr){5-7} \cmidrule(lr){8-10} \cmidrule(lr){11-13} \cmidrule(lr){14-16} \cmidrule(lr){17-19}
    & R@1 & R@5 & R@10 & R@1 & R@5 & R@10 & R@1 & R@5 & R@10 & R@1 & R@5 & R@10 & R@1 & R@5 & R@10 & R@1 & R@5 & R@10 \\
    \midrule
    CLIP4Clip-MeanP & 43.4 & 70.3 & 80.3 & 38.6 & 70.1 & 82.1 & 44.0 & 73.9 & 83.1 & 60.2 & 89.7 & 94.7 & 17.9 & 38.4 & 48.7 & 39.4 & 68.4 & 78.5 \\
    CLIP4Clip-seqTransf & 44.2 & 70.7 & 81.4 & 42.5 & 73.1 & 84.8 & 45.8 & 75.8 & 84.7 & 62.5 & 90.8 & 95.5 & 21.1 & 40.8 & 50.2 & 40.4 & 69.0 & 79.3 \\
    TeachCLIP w/o distill. & 44.6 & 71.8 & 81.2 & 42.9 & 73.8 & 85.4 & 46.0 & 75.7 & 84.4 & 63.3 & 91.1 & 95.7 & 21.6 & 40.3 & 48.9 & 43.3 & 70.8 & 79.3 \\
    X-CLIP & 46.7 & 72.7 & \textbf{82.9} & \textbf{44.6} & 74.2 & 85.1 & 46.1 & 76.0 & 85.0 & 62.3 & 90.5 & 95.3 & 21.9 & 41.8 & 51.0 & \textbf{45.4} & \textbf{73.7} & \textbf{82.2} \\
    X-Pool & 46.0 & 72.4 & 81.8 & 40.1 & 71.1 & 82.9 & 45.9 & 76.5 & 85.0 & 61.5 & 90.4 & 95.5 & 21.0 & 40.5 & 49.9 & 44.6 & 72.7 & \textbf{82.2} \\
    TS2-Net & 45.8 & 73.0 & 82.8 & 40.1 & 72.5 & 84.4 & 44.8 & 75.4 & 84.7 & 62.0 & 91.0 & 95.6 & 20.3 & 38.8 & 47.7 & 39.1 & 68.2 & 79.3 \\
    TeachCLIP & 45.6 & \textbf{73.1} & 82.4 & 43.1 & 74.2 & 85.3 & 46.9 & 76.7 & 85.4 & \textbf{63.7} & \textbf{91.5} & 95.9 & 21.1 & 41.4 & 50.5 & 44.5 & 71.0 & 80.6 \\
    \midrule
    \rowcolor{oursbg} Ours & 46.5 & 72.2 & 81.7 & 44.3 & \textbf{75.1} & \textbf{85.9} & 46.1 & 76.1 & 85.0 & 62.4 & 90.9 & 95.7 & \textbf{22.7} & \textbf{42.9} & 51.8 & 43.3 & 72.6 & 81.7 \\
    \rowcolor{oursbg} Ours + distillation & \textbf{47.0} & 72.5 & 82.1 & 43.9 & 74.8 & 85.5 & \textbf{47.1} & \textbf{77.3} & \textbf{85.7} & 63.3 & \textbf{91.5} & \textbf{96.0} & 22.5 & 42.4 & \textbf{52.2} & 44.6 & 72.4 & 81.8 \\
    \bottomrule
  \end{tabular}
  \label{tab:standard}
\end{table*}

\subsection{Setup}
We use ten benchmarks in two groups.
Four \emph{need dynamic content}, since their queries hinge on what changes over time (\cref{tab:specific}): SSv2-Template (SSv2-T) and SSv2-Label (SSv2-L)~\cite{goyal2017ssv2,lei2023singularity}, Charades~\cite{sigurdsson2016charades} and RTime~\cite{du2024rtime}.
SSv2-T queries replace the objects with ``something'', and SSv2-L queries name them.
The remaining six are \emph{standard benchmarks}, whose queries mostly describe appearance (\cref{tab:standard}): MSR-VTT~\cite{xu2016msrvtt} (1k-A split~\cite{yu2018jsfusion}), ActivityNet Captions~\cite{caba2015activitynet,krishna2017densecap}, MSVD~\cite{chen2011msvd}, VATEX~\cite{wang2019vatex} (HGR split~\cite{chen2020hgr}), LSMDC~\cite{rohrbach2017lsmdc} and DiDeMo~\cite{hendricks2017didemo}.
All methods use CLIP ViT-B/32 and are trained by us with the same frames: decoded at $1$ frame per second ($6$ on SSv2) and sampled uniformly, up to $12$ per clip (on ActivityNet and DiDeMo, all frames at $1$ fps, at most $64$).
On each benchmark, every method trains for the same number of epochs (\cref{sec:supp_setup}), except the original TeachCLIP code on MSR-VTT ($10$ epochs instead of $5$); X-Pool keeps its official settings.
Where a validation split exists (MSVD, VATEX, LSMDC, DiDeMo and RTime), we select the epoch on it; elsewhere we report the best test epoch, as is common practice~\cite{tian2024teachclip,luo2022clip4clip,ma2022xclip,liu2022ts2net,gorti2022xpool}. The rule is the same for every method, and every table entry is a mean over three seeds.

\paragraph{Baselines.}
Our main baselines are CLIP4Clip-seqTransf and TeachCLIP w/o distillation, the student of TeachCLIP~\cite{tian2024teachclip} trained without distillation (\cref{sec:seqtransf}).
TeachCLIP w/o distillation differs from our method in reading $x_t$ instead of $x_t-\bar{x}$, in scoring frames with an MLP instead of mean-queried attention, and in keeping no $\mu$ base and no null slot.
X-CLIP, X-Pool and TS2-Net use their official code or our re-implementation of it (published numbers: \cref{sec:supp_baselines}).

\paragraph{Distillation.}
TeachCLIP distills~\cite{hinton2015distilling} stronger teachers (X-CLIP, X-Pool and TS2-Net) into its student. To separate this effect from that of our method, we also train Ours + distillation with the coarse-grained loss of TeachCLIP, which matches the model's video--text similarities to those of the teachers.
TeachCLIP and Ours + distillation always use the same teachers: released TeachCLIP checkpoints on MSR-VTT, MSVD, DiDeMo and VATEX, all three on Charades and RTime, and X-CLIP elsewhere (\cref{sec:supp_distill}).

\paragraph{Cost.}
We time retrieval from the stored index to the top-10 list, without encoding the text, on one A40 GPU in fp16 (\cref{fig:teaser}b).
Over $100$k videos, a query takes $0.004$\,ms with one vector per clip and $3.9$ to $4.9$\,ms with X-Pool and X-CLIP, which compare the text with the frames at query time (formulas and other gallery sizes: \cref{sec:supp_cost}).

\subsection{Benchmarks that need dynamic content}
On SSv2 (\cref{tab:specific}), our method has the highest R@1 without distillation on both SSv2-T and SSv2-L ($51.5$ vs.\ $48.7$ for X-CLIP on SSv2-L), and with distillation on SSv2-T.
On Charades, whose videos contain several actions, it is on par with the best baseline, CLIP4Clip-seqTransf ($16.6$ vs.\ $16.9$; $17.1$ with four rounds chosen on test, \cref{sec:supp_rounds}).
On RTime, it ties the best baseline and is the best with distillation.  %

\subsection{Standard benchmarks}
On the standard benchmarks (\cref{tab:standard}), our method is the best on three of the six (LSMDC without distillation, MSR-VTT and MSVD with distillation).
Without distillation, it is within $0.2$ of X-CLIP on MSR-VTT.
On ActivityNet, five rounds bring it level with X-CLIP ($44.6$, chosen on test; \cref{sec:supp_rounds}).
With the same teachers, Ours + distillation is ahead of TeachCLIP on five of the six benchmarks; TeachCLIP leads only on VATEX ($+0.4$).  %
On MSR-VTT, the same distillation adds $+0.3$ to TeachCLIP w/o distillation and $+0.5$ to our method (\cref{sec:supp_distill}).
Full TeachCLIP, which also supervises the frame weights, reaches $45.6$, against $47.0$ for Ours + distillation.
The advantage thus comes from our method, not from distillation.  %
Across \cref{tab:specific,tab:standard}, taking the better of its results with and without distillation, our method is the best on six of the ten benchmarks and within $0.8$ R@1 of the best on the other four, with one vector per clip.
With CLIP ViT-B/16, our method stays above X-CLIP on SSv2-T ($83.7$ vs.\ $82.0$; \cref{sec:supp_backbone}).  %
\begin{table}[t]
  \caption{Component ablation (text-to-video R@1, mean over three seeds), one round ($R{=}1$). Each row modifies one part of our method; the last block changes one thing in TeachCLIP w/o distillation (seqTransf with AFA). $\Delta$: from unrounded means. More in \cref{sec:supp_ablation}.}
  \centering
  \small
  \setlength{\tabcolsep}{4pt}
  \renewcommand{\arraystretch}{0.92}  %
  \begin{tabular}{@{}lcccc@{}}
    \toprule
    & \multicolumn{2}{c}{SSv2-Template} & \multicolumn{2}{c}{MSR-VTT} \\
    \cmidrule(lr){2-3} \cmidrule(lr){4-5}
    Variant & R@1 & $\Delta$ & R@1 & $\Delta$ \\
    \midrule
    \rowcolor{oursbg} \textbf{Ours ($R{=}1$)} & 83.1 & -- & 46.0 & -- \\
    \midrule
    Candidates from $x_t$ (not $x_t-\bar{x}$) & 77.8 & $-5.4$ & 45.4 & $-0.6$ \\
    Learned fixed query (not $\mu$) & 82.8 & $-0.4$ & 45.2 & $-0.8$ \\  %
    No null slot & 80.1 & $-3.1$ & 45.2 & $-0.8$ \\
    \midrule
    TeachCLIP w/o distill. & 75.3 & $-7.9$ & 44.6 & $-1.4$ \\
    \quad + frame MLP & 75.9 & $-7.3$ & 44.3 & $-1.7$ \\  %
    \quad + input $x_t-\bar{x}$ & 82.2 & $-1.0$ & 44.8 & $-1.2$ \\  %
    \bottomrule
  \end{tabular}
  \label{tab:abl_parts}

  \vspace{8pt}
  \includegraphics[width=\linewidth]{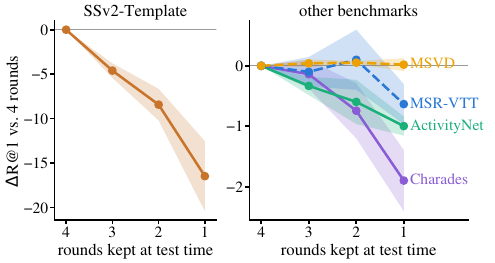}
  \captionof{figure}{\textbf{Dropping rounds at test time.} Change in R@1 when the later rounds of the four-round model are dropped without retraining. Band: mean$\pm$std over three seeds. Left: SSv2-Template; right: Charades and ActivityNet (solid), MSR-VTT and MSVD (dashed).}
  \label{fig:round_drop}
\end{table}

\subsection{Ablations}
\label{sec:ablation}
\Cref{tab:abl_parts} modifies one part of our method at a time, with one round; every modification lowers R@1 on both benchmarks.
On SSv2-T, the parts that extract the change matter most:
reading $x_t$ instead of $x_t-\bar{x}$ costs $5.4$, and removing the null slot $3.1$.
On MSR-VTT, keeping $\mu$ unchanged as the base matters most: every variant built on the clip vector of TeachCLIP w/o distillation, which keeps no $\mu$, loses $1.2$ to $1.7$.
A frame MLP with as many parameters helps little, so the gain does not come from capacity.
Feeding $x_t-\bar{x}$ to TeachCLIP w/o distillation recovers most of the gain on SSv2-T but not on MSR-VTT ($44.8$ vs.\ $46.0$). On SSv2-T it even exceeds our two-round model ($82.2$ vs.\ $81.6$), as $R$ is not tuned per benchmark.
A learned fixed attention query, the same for every clip, is close to ours on SSv2-T, which has few action types, but loses $0.8$ on MSR-VTT, whose content varies from clip to clip; there it helps that the attention query comes from the clip itself, as $\mu$ does.
To see whether the later rounds are used, we drop them at test time from the four-round model, without retraining (\cref{fig:round_drop}).
On MSVD and MSR-VTT, whose queries mostly describe static content, dropping rounds changes little (at most $-0.6$ with one round).
On SSv2-T, Charades and ActivityNet, which need dynamic content or contain several events, every dropped round costs R@1 ($-16.5$, $-1.9$ and $-1.0$ with one round).
The later rounds do not move to other frames: in the four-round model on SSv2-T, the frame weights of each round overlap by about $80\%$ with those of the previous round (the sum over frames of the smaller of the two weights).
What differs is how much they add: in the four-round model, each later round still adds an amount (\cref{eq:amount}) of $0.3$ to $0.4$ on SSv2-T, while on MSR-VTT the amount shrinks with every round.
So a model trained with several rounds refines the mean vector additively: where dynamic content matters, each round adds more of it from the same frames, and where the mean vector already describes the clip, the later rounds add little.
Training with more rounds, rather than dropping them, helps only on the longer videos.
On SSv2-T, one, two or four rounds give similar R@1 ($83.1$, $81.6$ and $83.9$; std $1.8$, $0.6$ and $3.5$). More rounds do not help beyond two on MSR-VTT and help slightly on Charades and ActivityNet (\cref{sec:supp_rounds}).

\begin{figure*}[t]
  \centering
  \includegraphics[width=\linewidth]{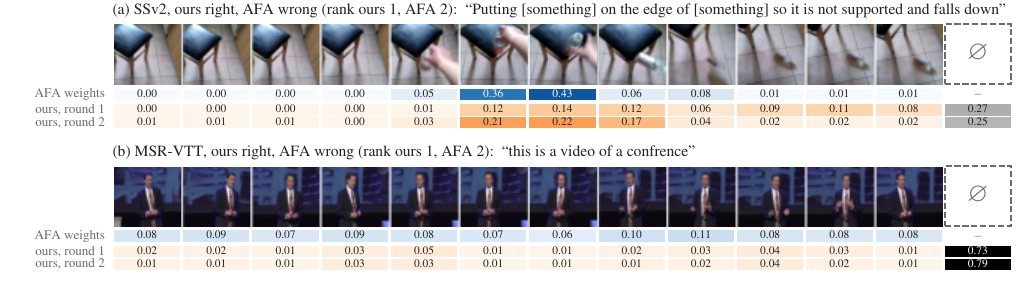}
  \caption{\textbf{Where AFA and our method put their weight.}
  Blue: the AFA weights of TeachCLIP w/o distillation over the 12 frames.
  Orange: the weights of our two rounds over the frames and the null slot ($\varnothing$); the weighted candidates are added to $\mu$.
  Queries are chosen by rule among those our method ranks first and AFA does not; the rule, counts and more examples, including a loss, are in \cref{sec:supp_examples}.
  In (a), AFA keeps only the moment the object is pushed, while our method adds that moment and the fall after it to $\mu$;
  in (b), most of the weight goes to the null slot, and $\mu$ is used almost as it is.}
  \label{fig:attn}
\end{figure*}

\section{Analysis}
\label{sec:why}
\paragraph{Retrieval is not frame selection.}
On SSv2-T, the AFA block of TeachCLIP w/o distillation puts $49\%$ of its weight on a single frame on average ($3.4$ effective frames out of $12$), while our method spreads it over $5.9$ to $6.7$ effective frames.
Both models look at the same moment: our method gives $20\%$ of its weight to the top frame of AFA and $26\%$ to the two frames next to it.
\Cref{fig:attn} shows the same pattern on single clips.
Unlike X-Pool, which weights the frames by their relevance to the text query (\cref{sec:selection}), our attention query is $\mu$: it collects the change that the mean vector misses, spread over the event and the frames around it.
\paragraph{How much is added.}
The null slot decides how much our method adds to the mean vector, separately for each clip.
On SSv2-T, whose queries ask about a change, the amount (\cref{eq:amount}) is $0.64$ in round 1 and $0.55$ in round 2; on MSR-VTT, where the mean vector already describes most clips, it is only $0.27$ and $0.18$ (two-round model, one seed).  %
This amount matters more than where the added content comes from.
We change the trained model only at test time (mean over three seeds).
If we spread the attention weight evenly over the frames but keep the learned amount, R@1 drops by $1.9$ on MSR-VTT and $1.3$ on SSv2-T.
If we also remove the null slot, so that the method always adds the full average of the candidates, R@1 drops by $18.3$ and $4.6$.
Using the mean vector of another clip as the attention query costs $0.8$ and $1.5$.

\paragraph{Sensitivity to temporal direction.}
\label{sec:direction}
Reversing the frames of the test clips lowers SSv2-T R@1 by about $40$ for four temporal models ($81.8\to38.3$ for ours, re-evaluated).
Our gain over TeachCLIP w/o distillation is $6.3$ on intact clips, $5.5$ on shuffled clips and $2.1$ on reversed clips.
On SSv2-L the dependence on order is even clearer (\cref{sec:supp_order}).
So part of the gain depends on the direction of time, not only on a better summary of the frames.

\begin{table}[t]
  \caption{\textbf{Cross-domain retrieval} (text-to-video R@1, mean over three seeds). Trained on one benchmark and tested on another; the checkpoint is chosen by the rule of the training benchmark, never on the target.
  MSR-VTT avg.\ of 5: DiDeMo, MSVD, VATEX, LSMDC and SSv2-Label; SSv2-T ($4.8$--$6.9$ for every model) is left out (per target: \cref{sec:supp_xdomain}).
  Gray: X-Pool, which pools the frames with the text query; bold: best among the other rows.
}
  \centering
  \footnotesize
  \setlength{\tabcolsep}{4pt}
  \renewcommand{\arraystretch}{0.92}  %
  \begin{tabular}{@{}lccc@{}}
    \toprule
    Train & MSR-VTT & SSv2-T & SSv2-L \\
    Test & avg.\ of 5 & MSR-VTT & MSR-VTT \\
    \midrule
    CLIP4Clip-MeanP & 27.7 & 15.9 & 25.8 \\
    CLIP4Clip-seqTransf & 30.7 & 15.3 & 24.9 \\
    TeachCLIP w/o distill. & 30.6 & 12.2 & 25.2 \\
    X-CLIP & 31.8 & 12.4 & 24.7 \\
    {\color{black!30}X-Pool} & {\color{black!30}34.8} & {\color{black!30}20.3} & {\color{black!30}29.7} \\
    TS2-Net & 29.1 & 13.0 & \textbf{26.7} \\
    TeachCLIP & 31.2 & 15.6 & 25.7 \\
    \midrule
    \rowcolor{oursbg} Ours & 31.9 & \textbf{17.5} & 25.6 \\
    \rowcolor{oursbg} Ours + distillation & \textbf{32.4} & 16.2 & 26.1 \\
    \bottomrule
  \end{tabular}
  \label{tab:xdomain}
\end{table}

\paragraph{Cross-domain generalization.}
\Cref{tab:xdomain} tests each model on benchmarks it was not trained on.
Trained on MSR-VTT, our method gives the best mean R@1 over five other benchmarks among the models other than X-Pool:
$31.9$, against $30.7$ for CLIP4Clip-seqTransf, $31.2$ for TeachCLIP and $31.8$ for X-CLIP.
Trained on SSv2-T, it keeps $17.5$ R@1 on MSR-VTT, while every baseline other than X-Pool falls below MeanP ($15.9$).
Trained on SSv2-L, TS2-Net transfers best and our method is in the middle.
Keeping $\mu$ as the base thus does not tie the method to the benchmark it was trained on.
X-Pool (gray in \cref{tab:xdomain}) transfers best overall but recomputes the video side for every text query.  %

\section{Conclusion and Limitations}
\label{sec:conclusion}
In a single-vector video-clip representation, the frame embeddings of a clip gather around their mean vector, and a temporal transformer captures the change only through corrections with almost the same direction for all frames.
We keep the mean vector as the base and let it query refinement candidates built from the deviations, with a null slot that lets it add little or nothing.
Each clip stays one vector, and retrieval costs the same as with mean pooling. Our method outperforms every baseline on Something-Something v2 and is the best or close to the best on the other benchmarks.

\paragraph{Limitations.}
Our method works best when the frames of a clip share one scene, so that their mean vector is a meaningful base.
In long videos with shot or scene changes, the mean vector mixes several scenes and is a weaker attention query for deciding what to add.
Even so, our method improves over TeachCLIP w/o distillation on ActivityNet ($44.3$ vs.\ $42.9$ R@1), whose videos last about two minutes.  %

On DiDeMo, whose queries describe several moments, X-CLIP stays ahead ($45.4$ vs.\ $43.3$).
On RTime, our method beats the best baseline only with distillation.

\FloatBarrier   %
\paragraph{Use of Generative AI} We used Claude Code (Anthropic) for coding, analysis and writing. The authors take responsibility for all content.

{
    \small
    \bibliographystyle{ieeenat_fullname}
    \bibliography{refs}

\begin{thebibliography}{37}
\providecommand{\natexlab}[1]{#1}
\providecommand{\url}[1]{\texttt{#1}}
\expandafter\ifx\csname urlstyle\endcsname\relax
  \providecommand{\doi}[1]{doi: #1}\else
  \providecommand{\doi}{doi: \begingroup \urlstyle{rm}\Url}\fi

\bibitem[Bagad et~al.(2023)Bagad, Tapaswi, and Snoek]{bagad2023testoftime}
Piyush Bagad, Makarand Tapaswi, and Cees G.~M. Snoek.
\newblock Test of time: Instilling video-language models with a sense of time.
\newblock In \emph{CVPR}, 2023.

\bibitem[Bondarenko et~al.(2023)Bondarenko, Nagel, and
  Blankevoort]{bondarenko2023quantizable}
Yelysei Bondarenko, Markus Nagel, and Tijmen Blankevoort.
\newblock Quantizable transformers: Removing outliers by helping attention
  heads do nothing.
\newblock In \emph{NeurIPS}, 2023.

\bibitem[Buch et~al.(2022)Buch, Eyzaguirre, Gaidon, Wu, Fei-Fei, and
  Niebles]{buch2022atp}
Shyamal Buch, Crist{\'o}bal Eyzaguirre, Adrien Gaidon, Jiajun Wu, Li Fei-Fei,
  and Juan~Carlos Niebles.
\newblock Revisiting the ``{Video}'' in video-language understanding.
\newblock In \emph{CVPR}, 2022.

\bibitem[Caba~Heilbron et~al.(2015)Caba~Heilbron, Escorcia, Ghanem, and
  Niebles]{caba2015activitynet}
Fabian Caba~Heilbron, Victor Escorcia, Bernard Ghanem, and Juan~Carlos Niebles.
\newblock {ActivityNet}: A large-scale video benchmark for human activity
  understanding.
\newblock In \emph{CVPR}, 2015.

\bibitem[Chen and Dolan(2011)]{chen2011msvd}
David~L. Chen and William~B. Dolan.
\newblock Collecting highly parallel data for paraphrase evaluation.
\newblock In \emph{ACL}, 2011.

\bibitem[Chen et~al.(2020)Chen, Zhao, Jin, and Wu]{chen2020hgr}
Shizhe Chen, Yida Zhao, Qin Jin, and Qi Wu.
\newblock Fine-grained video-text retrieval with hierarchical graph reasoning.
\newblock In \emph{CVPR}, 2020.

\bibitem[Darcet et~al.(2024)Darcet, Oquab, Mairal, and
  Bojanowski]{darcet2024registers}
Timoth{\'e}e Darcet, Maxime Oquab, Julien Mairal, and Piotr Bojanowski.
\newblock Vision transformers need registers.
\newblock In \emph{ICLR}, 2024.

\bibitem[Dosovitskiy et~al.(2021)Dosovitskiy, Beyer, Kolesnikov, Weissenborn,
  Zhai, Unterthiner, Dehghani, Minderer, Heigold, Gelly, Uszkoreit, and
  Houlsby]{dosovitskiy2021vit}
Alexey Dosovitskiy, Lucas Beyer, Alexander Kolesnikov, Dirk Weissenborn,
  Xiaohua Zhai, Thomas Unterthiner, Mostafa Dehghani, Matthias Minderer, Georg
  Heigold, Sylvain Gelly, Jakob Uszkoreit, and Neil Houlsby.
\newblock An image is worth 16x16 words: Transformers for image recognition at
  scale.
\newblock In \emph{ICLR}, 2021.

\bibitem[Du et~al.(2024)Du, Liu, and Jin]{du2024rtime}
Yang Du, Yuqi Liu, and Qin Jin.
\newblock Reversed in time: A novel temporal-emphasized benchmark for
  cross-modal video-text retrieval.
\newblock In \emph{ACM MM}, 2024.

\bibitem[Fang et~al.(2021)Fang, Xiong, Xu, and Chen]{fang2021clip2video}
Han Fang, Pengfei Xiong, Luhui Xu, and Yu Chen.
\newblock {CLIP2Video}: Mastering video-text retrieval via image {CLIP}.
\newblock \emph{arXiv preprint arXiv:2106.11097}, 2021.

\bibitem[Gorti et~al.(2022)Gorti, Vouitsis, Ma, Golestan, Volkovs, Garg, and
  Yu]{gorti2022xpool}
Satya~Krishna Gorti, No{\"e}l Vouitsis, Junwei Ma, Keyvan Golestan, Maksims
  Volkovs, Animesh Garg, and Guangwei Yu.
\newblock {X-Pool}: Cross-modal language-video attention for text-video
  retrieval.
\newblock In \emph{CVPR}, 2022.

\bibitem[Goyal et~al.(2017)Goyal, Ebrahimi~Kahou, Michalski, Materzy{\'n}ska,
  Westphal, Kim, Haenel, Fruend, Yianilos, Mueller-Freitag, Hoppe, Thurau, Bax,
  and Memisevic]{goyal2017ssv2}
Raghav Goyal, Samira Ebrahimi~Kahou, Vincent Michalski, Joanna Materzy{\'n}ska,
  Susanne Westphal, Heuna Kim, Valentin Haenel, Ingo Fruend, Peter Yianilos,
  Moritz Mueller-Freitag, Florian Hoppe, Christian Thurau, Ingo Bax, and Roland
  Memisevic.
\newblock The ``{Something Something}'' video database for learning and
  evaluating visual common sense.
\newblock In \emph{ICCV}, 2017.

\bibitem[Hendricks et~al.(2017)Hendricks, Wang, Shechtman, Sivic, Darrell, and
  Russell]{hendricks2017didemo}
Lisa~Anne Hendricks, Oliver Wang, Eli Shechtman, Josef Sivic, Trevor Darrell,
  and Bryan Russell.
\newblock Localizing moments in video with natural language.
\newblock In \emph{ICCV}, 2017.

\bibitem[Hinton et~al.(2015)Hinton, Vinyals, and Dean]{hinton2015distilling}
Geoffrey Hinton, Oriol Vinyals, and Jeff Dean.
\newblock Distilling the knowledge in a neural network.
\newblock \emph{arXiv preprint arXiv:1503.02531}, 2015.

\bibitem[Khattab and Zaharia(2020)]{khattab2020colbert}
Omar Khattab and Matei Zaharia.
\newblock {ColBERT}: Efficient and effective passage search via contextualized
  late interaction over {BERT}.
\newblock In \emph{SIGIR}, 2020.

\bibitem[Krishna et~al.(2017)Krishna, Hata, Ren, Fei-Fei, and
  Niebles]{krishna2017densecap}
Ranjay Krishna, Kenji Hata, Frederic Ren, Li Fei-Fei, and Juan~Carlos Niebles.
\newblock Dense-captioning events in videos.
\newblock In \emph{ICCV}, 2017.

\bibitem[Lei et~al.(2023)Lei, Berg, and Bansal]{lei2023singularity}
Jie Lei, Tamara~L. Berg, and Mohit Bansal.
\newblock Revealing single frame bias for video-and-language learning.
\newblock In \emph{ACL}, 2023.

\bibitem[Lin et~al.(2022)Lin, Geng, Zhang, Gao, de~Melo, Wang, Dai, Qiao, and
  Li]{lin2022evl}
Ziyi Lin, Shijie Geng, Renrui Zhang, Peng Gao, Gerard de Melo, Xiaogang Wang,
  Jifeng Dai, Yu Qiao, and Hongsheng Li.
\newblock Frozen {CLIP} models are efficient video learners.
\newblock In \emph{ECCV}, 2022.

\bibitem[Liu et~al.(2023)Liu, Huang, Li, Feng, Wu, and Li]{liu2023stan}
Ruyang Liu, Jingjia Huang, Ge Li, Jiashi Feng, Xinglong Wu, and Thomas~H. Li.
\newblock Revisiting temporal modeling for {CLIP}-based image-to-video
  knowledge transferring.
\newblock In \emph{CVPR}, 2023.

\bibitem[Liu et~al.(2022)Liu, Xiong, Xu, Cao, and Jin]{liu2022ts2net}
Yuqi Liu, Pengfei Xiong, Luhui Xu, Shengming Cao, and Qin Jin.
\newblock {TS2-Net}: Token shift and selection transformer for text-video
  retrieval.
\newblock In \emph{ECCV}, 2022.

\bibitem[Luo et~al.(2022)Luo, Ji, Zhong, Chen, Lei, Duan, and
  Li]{luo2022clip4clip}
Huaishao Luo, Lei Ji, Ming Zhong, Yang Chen, Wen Lei, Nan Duan, and Tianrui Li.
\newblock {CLIP4Clip}: An empirical study of {CLIP} for end to end video clip
  retrieval and captioning.
\newblock \emph{Neurocomputing}, 508:\penalty0 293--304, 2022.

\bibitem[Ma et~al.(2022)Ma, Xu, Sun, Yan, Zhang, and Ji]{ma2022xclip}
Yiwei Ma, Guohai Xu, Xiaoshuai Sun, Ming Yan, Ji Zhang, and Rongrong Ji.
\newblock {X-CLIP}: End-to-end multi-grained contrastive learning for
  video-text retrieval.
\newblock In \emph{ACM MM}, 2022.

\bibitem[Miller(2023)]{miller2023offbyone}
Evan Miller.
\newblock Attention is off by one.
\newblock \url{https://www.evanmiller.org/attention-is-off-by-one.html}, 2023.
\newblock Blog post.

\bibitem[{OpenAI}(2025)]{openai2025gptoss}
{OpenAI}.
\newblock gpt-oss-120b \& gpt-oss-20b model card.
\newblock \emph{arXiv preprint arXiv:2508.10925}, 2025.

\bibitem[Radford et~al.(2021)Radford, Kim, Hallacy, Ramesh, Goh, Agarwal,
  Sastry, Askell, Mishkin, Clark, Krueger, and Sutskever]{radford2021clip}
Alec Radford, Jong~Wook Kim, Chris Hallacy, Aditya Ramesh, Gabriel Goh,
  Sandhini Agarwal, Girish Sastry, Amanda Askell, Pamela Mishkin, Jack Clark,
  Gretchen Krueger, and Ilya Sutskever.
\newblock Learning transferable visual models from natural language
  supervision.
\newblock In \emph{ICML}, 2021.

\bibitem[Reddy et~al.(2025)Reddy, Martin, Yang, Yates, Sanders, Murray, Kriz,
  de~Melo, Van~Durme, and Chellappa]{reddy2025videocolbert}
Arun Reddy, Alexander Martin, Eugene Yang, Andrew Yates, Kate Sanders, Kenton
  Murray, Reno Kriz, Celso~M. de Melo, Benjamin Van~Durme, and Rama Chellappa.
\newblock {Video-ColBERT}: Contextualized late interaction for text-to-video
  retrieval.
\newblock In \emph{CVPR}, 2025.

\bibitem[Rohrbach et~al.(2017)Rohrbach, Torabi, Rohrbach, Tandon, Pal,
  Larochelle, Courville, and Schiele]{rohrbach2017lsmdc}
Anna Rohrbach, Atousa Torabi, Marcus Rohrbach, Niket Tandon, Christopher Pal,
  Hugo Larochelle, Aaron Courville, and Bernt Schiele.
\newblock Movie description.
\newblock \emph{IJCV}, 123:\penalty0 94--120, 2017.

\bibitem[Sigurdsson et~al.(2016)Sigurdsson, Varol, Wang, Farhadi, Laptev, and
  Gupta]{sigurdsson2016charades}
Gunnar~A. Sigurdsson, G{\"u}l Varol, Xiaolong Wang, Ali Farhadi, Ivan Laptev,
  and Abhinav Gupta.
\newblock Hollywood in homes: Crowdsourcing data collection for activity
  understanding.
\newblock In \emph{ECCV}, 2016.

\bibitem[Tian et~al.(2024)Tian, Zhao, Xin, Lan, and Li]{tian2024teachclip}
Kaibin Tian, Ruixiang Zhao, Zijie Xin, Bangxiang Lan, and Xirong Li.
\newblock Holistic features are almost sufficient for text-to-video retrieval.
\newblock In \emph{CVPR}, 2024.

\bibitem[van~den Oord et~al.(2018)van~den Oord, Li, and Vinyals]{oord2018cpc}
Aaron van~den Oord, Yazhe Li, and Oriol Vinyals.
\newblock Representation learning with contrastive predictive coding.
\newblock \emph{arXiv preprint arXiv:1807.03748}, 2018.

\bibitem[Vaswani et~al.(2017)Vaswani, Shazeer, Parmar, Uszkoreit, Jones, Gomez,
  Kaiser, and Polosukhin]{vaswani2017attention}
Ashish Vaswani, Noam Shazeer, Niki Parmar, Jakob Uszkoreit, Llion Jones,
  Aidan~N. Gomez, {\L}ukasz Kaiser, and Illia Polosukhin.
\newblock Attention is all you need.
\newblock In \emph{NeurIPS}, 2017.

\bibitem[Wang et~al.(2021)Wang, Tong, Ji, and Wu]{wang2021tdn}
Limin Wang, Zhan Tong, Bin Ji, and Gangshan Wu.
\newblock {TDN}: Temporal difference networks for efficient action recognition.
\newblock In \emph{CVPR}, 2021.

\bibitem[Wang et~al.(2019)Wang, Wu, Chen, Li, Wang, and Wang]{wang2019vatex}
Xin Wang, Jiawei Wu, Junkun Chen, Lei Li, Yuan-Fang Wang, and William~Yang
  Wang.
\newblock {VaTeX}: A large-scale, high-quality multilingual dataset for
  video-and-language research.
\newblock In \emph{ICCV}, 2019.

\bibitem[Xiao et~al.(2024)Xiao, Tian, Chen, Han, and
  Lewis]{xiao2024streamingllm}
Guangxuan Xiao, Yuandong Tian, Beidi Chen, Song Han, and Mike Lewis.
\newblock Efficient streaming language models with attention sinks.
\newblock In \emph{ICLR}, 2024.

\bibitem[Xu et~al.(2016)Xu, Mei, Yao, and Rui]{xu2016msrvtt}
Jun Xu, Tao Mei, Ting Yao, and Yong Rui.
\newblock {MSR-VTT}: A large video description dataset for bridging video and
  language.
\newblock In \emph{CVPR}, 2016.

\bibitem[Xue et~al.(2023)Xue, Sun, Liu, Fu, Song, Li, and Luo]{xue2023clipvip}
Hongwei Xue, Yuchong Sun, Bei Liu, Jianlong Fu, Ruihua Song, Houqiang Li, and
  Jiebo Luo.
\newblock {CLIP-ViP}: Adapting pre-trained image-text model to video-language
  alignment.
\newblock In \emph{ICLR}, 2023.

\bibitem[Yu et~al.(2018)Yu, Kim, and Kim]{yu2018jsfusion}
Youngjae Yu, Jongseok Kim, and Gunhee Kim.
\newblock A joint sequence fusion model for video question answering and
  retrieval.
\newblock In \emph{ECCV}, 2018.

\end{thebibliography}
}
\backrefsetup{disable}   %
\clearpage
\makeatletter\immediate\write\@auxout{\string\gdef\string\mainlastpage{\number\numexpr\value{page}-1\relax}}\makeatother
\appendix
\noindent{\Large\bfseries Appendix\par}\vspace{0.8em}

\providecommand{\suppstd}[1]{{\scriptsize$\pm$#1}}
\providecommand{\suppep}[1]{$_{#1}$}

\section{Implementation Details}
\label{sec:supp_setup}
This section gives the training and evaluation settings behind \cref{tab:specific,tab:standard} (\cref{sec:experiments}).

\paragraph{Code.}
Our method, CLIP4Clip-MeanP, CLIP4Clip-seqTransf, TeachCLIP w/o distillation and the distilled students run in one code base, which builds on the official TeachCLIP implementation~\cite{tian2024teachclip}.  %
The baselines that use other code are listed in \cref{sec:supp_baselines}.

\paragraph{Optimization.}
All models start from CLIP ViT-B/32 and fine-tune the whole network.
In our code, we use the BertAdam optimizer ($\beta_1{=}0.9$, $\beta_2{=}0.98$, $\epsilon{=}10^{-6}$) with weight decay $0.2$, gradient clipping at norm $1.0$, and a cosine schedule with linear warm-up over the first $10\%$ of the steps.  %
The learning rate is $10^{-4}$ for the new parameters and $10^{-7}$ for the CLIP encoders ($10^{-4}$ times $0.001$).  %
The MLP of the AFA block of TeachCLIP uses $10$ times the learning rate of the new parameters ($10^{-3}$).  %
In \cref{sec:experiments}, the official settings of X-Pool mean the following. On MSR-VTT, it uses the full official recipe. On the other benchmarks, it uses the learning rates and dropout of the official code, or the MSR-VTT values where the code gives none, with the batch sizes and epochs of \cref{tab:supp_setup} (\cref{sec:supp_baselines}).
The loss is the symmetric InfoNCE loss of CLIP4Clip over the batch (text-to-video and video-to-text, weight $1$ each).  %
The distillation losses and their weights are given in \cref{sec:supp_distill}.

\paragraph{Frames and text.}
Frames are decoded at $1$ frame per second ($6$ on SSv2) and sampled uniformly, up to $12$ per clip.
On ActivityNet and DiDeMo, we take all frames at $1$ fps, at most $64$, as in prior work (ActivityNet~\cite{luo2022clip4clip,ma2022xclip,liu2022ts2net}; DiDeMo~\cite{luo2022clip4clip}).  %
Captions are cut at $32$ tokens ($64$ on Charades, ActivityNet and DiDeMo).  %
\Cref{tab:supp_setup} lists the batch size and the number of epochs per benchmark.
Within a benchmark, every method uses the same values, with the exceptions in the table notes.

\begin{table}[t]
  \caption{Training settings per benchmark, shared by all methods. Frames: maximum number per clip. Words: maximum caption length in tokens.
  $^a$On MSR-VTT, TeachCLIP (original code) trains for $10$ epochs with batch $120$, and X-Pool (official recipe) uses batch $32$ (\cref{sec:supp_baselines}).}
  \centering
  \footnotesize
  \setlength{\tabcolsep}{3pt}
  \begin{tabular}{@{}lccccc@{}}
    \toprule
    Benchmark & Frames & Words & Batch & Epochs & Epoch chosen on \\
    \midrule
    SSv2-T / SSv2-L & 12 & 32 & 120 & 20 & test \\           %
    Charades        & 12 & 64 & 32  & 20 & test \\           %
    RTime           & 12 & 32 & 128 & 10 & validation \\     %
    \midrule
    MSR-VTT         & 12 & 32 & 128$^a$ & 5$^a$ & test \\    %
    ActivityNet     & 64 & 64 & 64  & 20 & test \\           %
    MSVD            & 12 & 32 & 128 & 5  & validation \\     %
    VATEX           & 12 & 32 & 128 & 5  & validation \\     %
    LSMDC           & 12 & 32 & 128 & 5  & validation \\     %
    DiDeMo          & 64 & 64 & 64  & 10 & validation \\     %
    \bottomrule
  \end{tabular}
  \label{tab:supp_setup}
\end{table}

\paragraph{Number of epochs.}
The number of epochs is fixed per benchmark and shared by all methods (\cref{tab:supp_setup}).  %
We checked that R@1 levels off by the last epoch (on the validation curve where a validation split exists, and on the test curve otherwise); this holds for $267$ of the $270$ runs behind \cref{tab:specific,tab:standard}.  %

\paragraph{Epoch selection.}
Where the benchmark has a validation split (MSVD, VATEX, LSMDC, DiDeMo and RTime), we pick the epoch with the best validation R@1 and report the test results of that epoch.  %
On RTime, the validation metric is Hard text-to-video R@1.  %
On the other benchmarks (SSv2-T, SSv2-L, Charades, MSR-VTT and ActivityNet), we report the epoch with the best test R@1, as in TeachCLIP~\cite{tian2024teachclip} and the public code of CLIP4Clip, X-CLIP, TS2-Net and X-Pool~\cite{luo2022clip4clip,ma2022xclip,liu2022ts2net,gorti2022xpool}.  %
R@5 and R@10 come from the same epoch as R@1, and ties go to the earliest epoch.  %
The rule is the same for every method.
One difference remains: X-Pool on MSR-VTT evaluates $10$ times per epoch, and we keep the best of these evaluations (\cref{sec:supp_baselines}).  %
The selected epoch of every run is listed in \cref{sec:supp_seeds}.

\paragraph{Protocols.}
We follow the official protocol of each benchmark where one exists.  %
SSv2-Template and SSv2-Label: the retrieval split of Singularity~\cite{lei2023singularity}, with $168{,}913$ training and $2{,}088$ test videos.  %
Charades has no official retrieval protocol; we use the $7{,}985$ training and $1{,}863$ test videos of the official split, with the scripts as queries, as in prior retrieval work.  %
RTime (Hard)~\cite{du2024rtime}: $11{,}771$ training, $500$ validation and $1{,}000$ test source videos in the official split, of which $990$ test videos are available (the RTime paper also counts the reversed copies: $18{,}537$, $1{,}000$ and $2{,}000$ videos).  %
The forward and the reversed captions are both queries, and the gallery holds every video and its reversed copy.  %
MSR-VTT: we train on the 9k training split and test on the 1k-A split~\cite{yu2018jsfusion}.  %
ActivityNet: $10{,}009$ training videos and the $4{,}917$ videos of val\_1 as the test set, with the captions of each video concatenated.  %
MSVD: $1{,}200$ training, $100$ validation and $670$ test videos ($48{,}774$ training and $27{,}763$ test captions).  %
VATEX: the HGR split~\cite{chen2020hgr} ($25{,}991$ training, $1{,}500$ validation and $1{,}500$ test videos); $22{,}095$ of the training and $1{,}303$ of the test videos are still available, and all methods use these.  %
LSMDC: the LSMDC16 training set ($101{,}045$ videos available) and the $1{,}000$-clip public test set.  %
DiDeMo: a test set of $1{,}004$ videos, as in CLIP4Clip~\cite{luo2022clip4clip}.  %

\paragraph{Seeds.}
Every cell of \cref{tab:specific,tab:standard} is the mean over three seeds.
The seed also sets the order of the training data.  %

\section{Retrained Baselines and Published Numbers}
\label{sec:supp_baselines}
We train every baseline ourselves with the settings of \cref{sec:supp_setup}, except for the official recipes listed below; we do not tune any baseline per benchmark.

\paragraph{Code of each baseline.}
\begin{itemize}
  \item \textbf{CLIP4Clip-MeanP, CLIP4Clip-seqTransf and TeachCLIP w/o distillation}: our code (\cref{sec:supp_setup}).
  \item \textbf{X-CLIP}: on MSR-VTT, the official code, with batch $128$ and $5$ epochs.  %
  On the other benchmarks, our re-implementation in our code, trained with our recipe. On MSR-VTT, we report the official code because it is closer to the published number ($46.7$ vs.\ $46.1$); our re-implementation reaches $47.2$ there (best test epoch, three seeds).  %
  \item \textbf{X-Pool}: on MSR-VTT, our re-implementation trained with the official X-Pool recipe: batch $32$, learning rate $3\times10^{-5}$ ($10^{-6}$ for CLIP), $5$ epochs and $10$ evaluations per epoch.  %
  The released X-Pool checkpoint gives the same R@1 in our evaluation ($46.0$).  %
  On the other benchmarks, our re-implementation with our batch sizes and epochs, the learning rates of the official X-Pool code ($10^{-5}$, and $10^{-6}$ for CLIP, on MSVD and LSMDC; elsewhere, where the code gives no setting, the MSR-VTT values $3\times10^{-5}$ and $10^{-6}$) and its dropout ($0.4$ on MSVD, $0.3$ elsewhere).  %
  \item \textbf{TS2-Net}: our re-implementation with token shift on all benchmarks, trained with our recipe.  %
  On MSR-VTT, the published setting (12 frames, batch $128$) is the same as our setting.  %
  \item \textbf{TeachCLIP}: on MSR-VTT, the original code with its released configuration ($10$ epochs, batch $120$) and the three teachers it releases (X-CLIP, TS2-Net and X-Pool).  %
  On the other benchmarks, its student and losses in our code, with the same teachers as Ours + distillation (\cref{sec:supp_distill}).  %
\end{itemize}

\paragraph{Comparison with published numbers.}
\Cref{tab:supp_baselines} compares our numbers with those that each baseline reports in its own paper, with CLIP ViT-B/32 and no test-time post-processing.  %
Our numbers are the means of \cref{tab:specific,tab:standard}.
The settings differ as follows.
(i) On VATEX, we test on the $1{,}303$ available test videos, not on all $1{,}500$.  %
(ii) On DiDeMo, we use batch $64$, since batch $128$ runs out of memory.  %
(iii) On MSVD, VATEX, LSMDC and DiDeMo, we select the epoch on the validation split (\cref{sec:supp_setup}).
(iv) The SSv2 numbers of CLIP4Clip-seqTransf were obtained by the Singularity authors, who do not state the ViT size~\cite{lei2023singularity}.  %
(v) For TeachCLIP outside MSR-VTT, the teachers differ from those of the TeachCLIP paper (\cref{sec:supp_distill}).  %
On MSR-VTT, where the setting is closest, our numbers are within $1.2$ R@1 of the published ones for every baseline.  %
Charades and RTime are not in the table: for Charades, we found only re-implementations of these baselines in other papers, and the RTime paper reports ViT-B/16 models only.  %

\begin{table*}[t]
  \caption{Text-to-video R@1 of the baselines: numbers published in each method's own paper (CLIP ViT-B/32, no test-time post-processing) and reproduced (our runs under the setting of \cref{sec:supp_setup}; the means of \cref{tab:specific,tab:standard}). ``--'': not reported.
  $^\ast$Published: the HGR test set ($1{,}500$ videos where the paper states it); reproduced: the $1{,}303$ available test videos.
  $^\dagger$Run by the Singularity authors~\cite{lei2023singularity}; ViT size not stated.}
  \label{tab:supp_baselines}
  \centering
  \small
  \setlength{\tabcolsep}{4.5pt}
  \begin{tabular}{llcccccccc}
    \toprule
    Method & Numbers & MSR-VTT & ActivityNet & MSVD & VATEX$^\ast$ & LSMDC & DiDeMo & SSv2-T & SSv2-L \\
    \midrule
    CLIP4Clip-MeanP     & published & 43.1 & 40.5 & 46.2 & --   & 20.7 & 43.4 & --   & --   \\  %
                        & reproduced & 43.4 & 38.6 & 44.0 & 60.2 & 17.9 & 39.4 & 51.9 & 39.7 \\  %
    \midrule
    CLIP4Clip-seqTransf & published & 44.5 & 40.5 & 45.2 & --   & 22.6 & 42.8 & 77.0$^\dagger$ & 43.1$^\dagger$ \\  %
                        & reproduced & 44.2 & 42.5 & 45.8 & 62.5 & 21.1 & 40.4 & 78.5 & 48.2 \\  %
    \midrule
    X-CLIP              & published & 46.1 & 44.3 & 47.1 & --   & 23.3 & 45.2 & --   & --   \\  %
                        & reproduced & 46.7 & 44.6 & 46.1 & 62.3 & 21.9 & 45.4 & 80.3 & 48.7 \\  %
    \midrule
    X-Pool              & published & 46.9 & --   & 47.2 & --   & 25.2 & --   & --   & --   \\  %
                        & reproduced & 46.0 & 40.1 & 45.9 & 61.5 & 21.0 & 44.6 & 55.9 & 38.2 \\  %
    \midrule
    TS2-Net             & published & 47.0 & 41.0 & --   & 59.1 & 23.4 & 41.8 & --   & --   \\  %
                        & reproduced & 45.8 & 40.1 & 44.8 & 62.0 & 20.3 & 39.1 & 69.1 & 46.3 \\  %
    \midrule
    TeachCLIP           & published & 46.8 & 42.2 & 47.4 & 63.6 & --   & 43.7 & --   & --   \\  %
                        & reproduced & 45.6 & 43.1 & 46.9 & 63.7 & 21.1 & 44.5 & 78.9 & 48.4 \\  %
    \bottomrule
  \end{tabular}
\end{table*}

\section{Per-Seed Results}
\label{sec:supp_seeds}
\Cref{tab:supp_seeds_a,tab:supp_seeds_b,tab:supp_seeds_c,tab:supp_seeds_d,tab:supp_seeds_e} give the mean and sample standard deviation of R@1, R@5 and R@10 over three seeds for every cell of \cref{tab:specific,tab:standard}.
They also give the R@1 of each seed and the epoch selected for it (\cref{sec:supp_setup}).
Epochs are counted from $1$.
The means, and every difference in this appendix, are computed from the unrounded per-seed values, so they can differ by $0.1$ from values computed from the rounded numbers shown.
SSv2-Template has only $174$ test queries, so one query changes R@1 by $0.57$; its standard deviations are the largest (up to $4.27$ R@1 for X-CLIP).  %
Elsewhere, the standard deviation of R@1 is at most $1.27$ (TS2-Net on DiDeMo).  %

\begin{table*}[t]
  \caption{Per-seed results on SSv2-Template and SSv2-Label (\cref{tab:specific}). Left of each block: mean $\pm$ sample standard deviation over three seeds. Right: R@1 of each seed, with the selected epoch (1-based) as a subscript. Epoch chosen on the test set; $20$ epochs.}
  \label{tab:supp_seeds_a}
  \centering
  \footnotesize
  \setlength{\tabcolsep}{2.5pt}
  \begin{tabular}{lccc@{\hspace{6pt}}ccc@{\hspace{10pt}}ccc@{\hspace{6pt}}ccc}
    \toprule
    Method & \multicolumn{6}{c}{SSv2-Template} & \multicolumn{6}{c}{SSv2-Label} \\
    \cmidrule(lr){2-7} \cmidrule(lr){8-13}
    & \multicolumn{3}{c}{mean $\pm$ std} & \multicolumn{3}{c}{R@1 per seed (epoch)} & \multicolumn{3}{c}{mean $\pm$ std} & \multicolumn{3}{c}{R@1 per seed (epoch)} \\
    \cmidrule(lr){2-4} \cmidrule(lr){5-7} \cmidrule(lr){8-10} \cmidrule(lr){11-13}
    & R@1 & R@5 & R@10 & 1 & 2 & 3 & R@1 & R@5 & R@10 & 1 & 2 & 3 \\
    \midrule
    CLIP4Clip-MeanP & 51.9\suppstd{0.33} & 87.6\suppstd{1.20} & 96.2\suppstd{0.33} & 51.7\suppep{10} & 52.3\suppep{13} & 51.7\suppep{14} & 39.7\suppstd{0.26} & 67.1\suppstd{0.50} & 76.5\suppstd{0.56} & 39.5\suppep{18} & 40.0\suppep{14} & 39.5\suppep{15} \\  %
    CLIP4Clip-seqTransf & 78.5\suppstd{2.39} & 97.5\suppstd{0.33} & 99.4\suppstd{0.00} & 80.5\suppep{15} & 79.3\suppep{16} & 75.9\suppep{16} & 48.2\suppstd{0.31} & 74.8\suppstd{0.09} & 82.5\suppstd{0.28} & 48.2\suppep{19} & 48.5\suppep{20} & 47.9\suppep{20} \\  %
    TeachCLIP w/o distill. & 75.3\suppstd{1.15} & 97.9\suppstd{0.66} & 99.2\suppstd{0.33} & 74.1\suppep{18} & 75.3\suppep{14} & 76.4\suppep{18} & 48.0\suppstd{0.34} & 75.2\suppstd{0.72} & 83.4\suppstd{0.23} & 47.6\suppep{19} & 48.1\suppep{17} & 48.3\suppep{16} \\  %
    X-CLIP & 80.3\suppstd{4.27} & 98.3\suppstd{0.60} & 99.4\suppstd{0.00} & 85.1\suppep{16} & 78.7\suppep{17} & 77.0\suppep{13} & 48.7\suppstd{1.25} & 74.8\suppstd{0.51} & 83.1\suppstd{0.46} & 49.1\suppep{15} & 47.3\suppep{18} & 49.7\suppep{13} \\  %
    X-Pool & 55.9\suppstd{3.26} & 91.6\suppstd{0.35} & 97.1\suppstd{0.55} & 52.3\suppep{9} & 58.6\suppep{7} & 56.9\suppep{9} & 38.2\suppstd{0.26} & 66.9\suppstd{0.21} & 77.1\suppstd{0.25} & 37.9\suppep{10} & 38.4\suppep{15} & 38.3\suppep{12} \\  %
    TS2-Net & 69.1\suppstd{3.46} & 95.2\suppstd{0.35} & 98.5\suppstd{0.35} & 72.4\suppep{16} & 69.5\suppep{9} & 65.5\suppep{17} & 46.3\suppstd{0.50} & 74.6\suppstd{0.36} & 82.3\suppstd{0.38} & 46.8\suppep{20} & 45.8\suppep{19} & 46.3\suppep{18} \\  %
    TeachCLIP & 78.9\suppstd{1.84} & 98.3\suppstd{1.15} & 99.2\suppstd{0.33} & 77.6\suppep{3} & 78.2\suppep{15} & 81.0\suppep{16} & 48.4\suppstd{0.59} & 75.2\suppstd{0.58} & 83.4\suppstd{0.71} & 49.1\suppep{20} & 47.9\suppep{13} & 48.3\suppep{20} \\  %
    \midrule
    \rowcolor{oursbg} Ours & 81.6\suppstd{0.58} & 97.9\suppstd{0.88} & 99.6\suppstd{0.33} & 81.0\suppep{16} & 81.6\suppep{13} & 82.2\suppep{13} & 51.5\suppstd{0.49} & 76.3\suppstd{0.34} & 84.0\suppstd{0.47} & 51.0\suppep{18} & 52.0\suppep{19} & 51.4\suppep{17} \\  %
    \rowcolor{oursbg} Ours + distillation & 82.2\suppstd{2.30} & 97.9\suppstd{0.33} & 99.4\suppstd{0.58} & 84.5\suppep{18} & 82.2\suppep{15} & 79.9\suppep{13} & 50.7\suppstd{0.49} & 77.1\suppstd{0.21} & 84.9\suppstd{0.76} & 51.1\suppep{20} & 50.9\suppep{18} & 50.2\suppep{20} \\  %
    \bottomrule
  \end{tabular}
\end{table*}

\begin{table*}[t]
  \caption{Per-seed results on Charades and RTime (Hard) (\cref{tab:specific}); format as in \cref{tab:supp_seeds_a}. Charades: epoch chosen on the test set, $20$ epochs; Ours uses $R{=}2$ rounds as in \cref{tab:specific}. RTime: epoch chosen on the validation split, $10$ epochs.}
  \label{tab:supp_seeds_b}
  \centering
  \footnotesize
  \setlength{\tabcolsep}{2.5pt}
  \begin{tabular}{lccc@{\hspace{6pt}}ccc@{\hspace{10pt}}ccc@{\hspace{6pt}}ccc}
    \toprule
    Method & \multicolumn{6}{c}{Charades} & \multicolumn{6}{c}{RTime (Hard)} \\
    \cmidrule(lr){2-7} \cmidrule(lr){8-13}
    & \multicolumn{3}{c}{mean $\pm$ std} & \multicolumn{3}{c}{R@1 per seed (epoch)} & \multicolumn{3}{c}{mean $\pm$ std} & \multicolumn{3}{c}{R@1 per seed (epoch)} \\
    \cmidrule(lr){2-4} \cmidrule(lr){5-7} \cmidrule(lr){8-10} \cmidrule(lr){11-13}
    & R@1 & R@5 & R@10 & 1 & 2 & 3 & R@1 & R@5 & R@10 & 1 & 2 & 3 \\
    \midrule
    CLIP4Clip-MeanP & 15.0\suppstd{0.36} & 32.3\suppstd{0.87} & 41.7\suppstd{0.58} & 14.9\suppep{10} & 15.4\suppep{16} & 14.7\suppep{8} & 34.8\suppstd{0.21} & 84.4\suppstd{0.86} & 91.4\suppstd{0.37} & 34.7\suppep{10} & 35.0\suppep{4} & 34.6\suppep{10} \\  %
    CLIP4Clip-seqTransf & 16.9\suppstd{0.42} & 36.2\suppstd{0.17} & 46.9\suppstd{0.44} & 16.8\suppep{20} & 17.4\suppep{16} & 16.6\suppep{18} & 35.1\suppstd{0.15} & 83.9\suppstd{0.47} & 91.8\suppstd{0.37} & 35.3\suppep{8} & 35.2\suppep{8} & 34.9\suppep{6} \\  %
    TeachCLIP w/o distill. & 16.6\suppstd{0.40} & 35.6\suppstd{0.90} & 46.2\suppstd{1.51} & 17.0\suppep{18} & 16.2\suppep{13} & 16.5\suppep{18} & 35.5\suppstd{1.22} & 84.5\suppstd{1.03} & 91.8\suppstd{1.02} & 36.8\suppep{9} & 34.4\suppep{10} & 35.4\suppep{6} \\  %
    X-CLIP & 15.7\suppstd{0.90} & 34.3\suppstd{0.55} & 44.2\suppstd{0.76} & 15.8\suppep{19} & 14.8\suppep{15} & 16.6\suppep{14} & 34.1\suppstd{0.25} & 81.6\suppstd{0.35} & 89.7\suppstd{0.40} & 34.1\suppep{5} & 34.4\suppep{8} & 33.9\suppep{3} \\  %
    X-Pool & 12.1\suppstd{0.15} & 29.0\suppstd{0.71} & 39.4\suppstd{0.26} & 12.2\suppep{12} & 11.9\suppep{7} & 12.1\suppep{4} & 35.2\suppstd{0.25} & 85.7\suppstd{0.23} & 92.4\suppstd{0.66} & 34.9\suppep{4} & 35.4\suppep{1} & 35.2\suppep{7} \\  %
    TS2-Net & 14.3\suppstd{0.51} & 32.8\suppstd{0.81} & 42.5\suppstd{0.80} & 14.7\suppep{17} & 13.7\suppep{14} & 14.4\suppep{15} & 34.4\suppstd{0.40} & 83.7\suppstd{0.74} & 91.3\suppstd{0.49} & 34.8\suppep{9} & 34.0\suppep{6} & 34.3\suppep{6} \\  %
    TeachCLIP & 15.5\suppstd{0.66} & 34.1\suppstd{0.95} & 43.8\suppstd{1.34} & 15.4\suppep{12} & 14.9\suppep{11} & 16.2\suppep{12} & 32.5\suppstd{1.05} & 82.0\suppstd{1.76} & 89.9\suppstd{1.40} & 33.3\suppep{3} & 31.3\suppep{6} & 32.8\suppep{3} \\  %
    \midrule
    \rowcolor{oursbg} Ours & 16.6\suppstd{0.58} & 35.8\suppstd{0.72} & 46.4\suppstd{1.18} & 16.3\suppep{16} & 16.3\suppep{17} & 17.3\suppep{12} & 35.5\suppstd{1.08} & 84.7\suppstd{0.35} & 91.8\suppstd{0.03} & 36.7\suppep{10} & 35.1\suppep{6} & 34.7\suppep{3} \\  %
    \rowcolor{oursbg} Ours + distillation & 16.3\suppstd{0.31} & 34.9\suppstd{0.57} & 45.3\suppstd{1.01} & 16.4\suppep{15} & 16.6\suppep{19} & 16.0\suppep{13} & 36.3\suppstd{0.33} & 85.6\suppstd{0.97} & 92.5\suppstd{0.85} & 36.0\suppep{8} & 36.3\suppep{10} & 36.7\suppep{10} \\  %
    \bottomrule
  \end{tabular}
\end{table*}

\begin{table*}[t]
  \caption{Per-seed results on MSR-VTT and ActivityNet (\cref{tab:standard}); format as in \cref{tab:supp_seeds_a}. Epoch chosen on the test set. MSR-VTT: $5$ epochs ($10$ for TeachCLIP, original code); X-Pool keeps the best of $10$ evaluations per epoch, and its subscript is the epoch of that evaluation. ActivityNet: $20$ epochs.}
  \label{tab:supp_seeds_c}
  \centering
  \footnotesize
  \setlength{\tabcolsep}{2.5pt}
  \begin{tabular}{lccc@{\hspace{6pt}}ccc@{\hspace{10pt}}ccc@{\hspace{6pt}}ccc}
    \toprule
    Method & \multicolumn{6}{c}{MSR-VTT} & \multicolumn{6}{c}{ActivityNet} \\
    \cmidrule(lr){2-7} \cmidrule(lr){8-13}
    & \multicolumn{3}{c}{mean $\pm$ std} & \multicolumn{3}{c}{R@1 per seed (epoch)} & \multicolumn{3}{c}{mean $\pm$ std} & \multicolumn{3}{c}{R@1 per seed (epoch)} \\
    \cmidrule(lr){2-4} \cmidrule(lr){5-7} \cmidrule(lr){8-10} \cmidrule(lr){11-13}
    & R@1 & R@5 & R@10 & 1 & 2 & 3 & R@1 & R@5 & R@10 & 1 & 2 & 3 \\
    \midrule
    CLIP4Clip-MeanP & 43.4\suppstd{0.61} & 70.3\suppstd{0.31} & 80.3\suppstd{0.10} & 43.1\suppep{2} & 43.0\suppep{2} & 44.1\suppep{2} & 38.6\suppstd{0.15} & 70.1\suppstd{0.15} & 82.1\suppstd{0.10} & 38.6\suppep{14} & 38.8\suppep{14} & 38.5\suppep{12} \\  %
    CLIP4Clip-seqTransf & 44.2\suppstd{0.31} & 70.7\suppstd{0.72} & 81.4\suppstd{0.50} & 44.3\suppep{2} & 43.9\suppep{3} & 44.5\suppep{3} & 42.5\suppstd{0.06} & 73.1\suppstd{0.10} & 84.8\suppstd{0.12} & 42.5\suppep{16} & 42.4\suppep{16} & 42.5\suppep{18} \\  %
    TeachCLIP w/o distill. & 44.6\suppstd{0.21} & 71.8\suppstd{0.45} & 81.2\suppstd{0.66} & 44.7\suppep{2} & 44.8\suppep{4} & 44.4\suppep{3} & 42.9\suppstd{0.32} & 73.8\suppstd{0.15} & 85.4\suppstd{0.26} & 43.3\suppep{17} & 42.8\suppep{17} & 42.7\suppep{18} \\  %
    X-CLIP & 46.7\suppstd{0.30} & 72.7\suppstd{0.25} & 82.9\suppstd{0.81} & 46.4\suppep{3} & 47.0\suppep{3} & 46.7\suppep{2} & 44.6\suppstd{0.31} & 74.2\suppstd{0.26} & 85.1\suppstd{0.06} & 44.9\suppep{17} & 44.3\suppep{19} & 44.5\suppep{17} \\  %
    X-Pool & 46.0\suppstd{0.06} & 72.4\suppstd{0.44} & 81.8\suppstd{0.82} & 46.0\suppep{3} & 46.1\suppep{3} & 46.0\suppep{4} & 40.1\suppstd{0.23} & 71.1\suppstd{0.26} & 82.9\suppstd{0.31} & 40.2\suppep{15} & 40.2\suppep{15} & 39.8\suppep{15} \\  %
    TS2-Net & 45.8\suppstd{0.50} & 73.0\suppstd{0.15} & 82.8\suppstd{0.15} & 45.3\suppep{4} & 46.3\suppep{2} & 45.9\suppep{3} & 40.1\suppstd{0.32} & 72.5\suppstd{0.44} & 84.4\suppstd{0.12} & 39.7\suppep{18} & 40.2\suppep{18} & 40.3\suppep{18} \\  %
    TeachCLIP & 45.6\suppstd{0.35} & 73.1\suppstd{0.60} & 82.4\suppstd{0.61} & 45.6\suppep{3} & 45.9\suppep{7} & 45.2\suppep{4} & 43.1\suppstd{0.56} & 74.2\suppstd{0.35} & 85.3\suppstd{0.15} & 42.5\suppep{20} & 43.2\suppep{18} & 43.6\suppep{17} \\  %
    \midrule
    \rowcolor{oursbg} Ours & 46.5\suppstd{0.40} & 72.2\suppstd{0.75} & 81.7\suppstd{0.76} & 46.1\suppep{3} & 46.6\suppep{2} & 46.9\suppep{3} & 44.3\suppstd{0.12} & 75.1\suppstd{0.35} & 85.9\suppstd{0.10} & 44.2\suppep{17} & 44.4\suppep{17} & 44.4\suppep{19} \\  %
    \rowcolor{oursbg} Ours + distillation & 47.0\suppstd{0.57} & 72.5\suppstd{0.12} & 82.1\suppstd{0.58} & 47.5\suppep{3} & 46.4\suppep{2} & 47.2\suppep{3} & 43.9\suppstd{0.59} & 74.8\suppstd{0.10} & 85.5\suppstd{0.15} & 44.3\suppep{16} & 43.2\suppep{17} & 44.1\suppep{17} \\  %
    \bottomrule
  \end{tabular}
\end{table*}

\begin{table*}[t]
  \caption{Per-seed results on MSVD and VATEX (\cref{tab:standard}); format as in \cref{tab:supp_seeds_a}. Epoch chosen on the validation split; $5$ epochs. $^\ast$VATEX: the $1{,}303$ available test videos of the HGR split.}
  \label{tab:supp_seeds_d}
  \centering
  \footnotesize
  \setlength{\tabcolsep}{2.5pt}
  \begin{tabular}{lccc@{\hspace{6pt}}ccc@{\hspace{10pt}}ccc@{\hspace{6pt}}ccc}
    \toprule
    Method & \multicolumn{6}{c}{MSVD} & \multicolumn{6}{c}{VATEX$^\ast$} \\
    \cmidrule(lr){2-7} \cmidrule(lr){8-13}
    & \multicolumn{3}{c}{mean $\pm$ std} & \multicolumn{3}{c}{R@1 per seed (epoch)} & \multicolumn{3}{c}{mean $\pm$ std} & \multicolumn{3}{c}{R@1 per seed (epoch)} \\
    \cmidrule(lr){2-4} \cmidrule(lr){5-7} \cmidrule(lr){8-10} \cmidrule(lr){11-13}
    & R@1 & R@5 & R@10 & 1 & 2 & 3 & R@1 & R@5 & R@10 & 1 & 2 & 3 \\
    \midrule
    CLIP4Clip-MeanP & 44.0\suppstd{0.56} & 73.9\suppstd{0.18} & 83.1\suppstd{0.17} & 43.5\suppep{1} & 44.6\suppep{2} & 43.9\suppep{3} & 60.2\suppstd{0.25} & 89.7\suppstd{0.10} & 94.7\suppstd{0.00} & 60.0\suppep{4} & 60.5\suppep{3} & 60.2\suppep{4} \\  %
    CLIP4Clip-seqTransf & 45.8\suppstd{0.82} & 75.8\suppstd{0.60} & 84.7\suppstd{0.28} & 44.9\suppep{1} & 46.3\suppep{5} & 46.4\suppep{3} & 62.5\suppstd{0.31} & 90.8\suppstd{0.20} & 95.5\suppstd{0.15} & 62.2\suppep{4} & 62.8\suppep{4} & 62.6\suppep{4} \\  %
    TeachCLIP w/o distill. & 46.0\suppstd{0.17} & 75.7\suppstd{0.16} & 84.4\suppstd{0.12} & 46.2\suppep{2} & 46.0\suppep{1} & 45.9\suppep{1} & 63.3\suppstd{0.06} & 91.1\suppstd{0.32} & 95.7\suppstd{0.15} & 63.3\suppep{5} & 63.3\suppep{5} & 63.2\suppep{5} \\  %
    X-CLIP & 46.1\suppstd{0.44} & 76.0\suppstd{0.20} & 85.0\suppstd{0.15} & 45.9\suppep{1} & 45.8\suppep{3} & 46.6\suppep{3} & 62.3\suppstd{0.20} & 90.5\suppstd{0.06} & 95.3\suppstd{0.12} & 62.1\suppep{4} & 62.3\suppep{5} & 62.5\suppep{4} \\  %
    X-Pool & 45.9\suppstd{0.15} & 76.5\suppstd{0.10} & 85.0\suppstd{0.06} & 45.9\suppep{5} & 45.8\suppep{4} & 46.1\suppep{3} & 61.5\suppstd{0.32} & 90.4\suppstd{0.15} & 95.5\suppstd{0.06} & 61.3\suppep{2} & 61.4\suppep{2} & 61.9\suppep{5} \\  %
    TS2-Net & 44.8\suppstd{0.26} & 75.4\suppstd{0.46} & 84.7\suppstd{0.26} & 44.6\suppep{2} & 44.7\suppep{2} & 45.1\suppep{3} & 62.0\suppstd{0.17} & 91.0\suppstd{0.06} & 95.6\suppstd{0.15} & 61.9\suppep{5} & 61.9\suppep{5} & 62.2\suppep{4} \\  %
    TeachCLIP & 46.9\suppstd{0.92} & 76.7\suppstd{0.89} & 85.4\suppstd{0.60} & 47.2\suppep{5} & 45.8\suppep{1} & 47.5\suppep{3} & 63.7\suppstd{0.12} & 91.5\suppstd{0.10} & 95.9\suppstd{0.12} & 63.8\suppep{5} & 63.6\suppep{5} & 63.6\suppep{4} \\  %
    \midrule
    \rowcolor{oursbg} Ours & 46.1\suppstd{0.52} & 76.1\suppstd{0.73} & 85.0\suppstd{0.53} & 45.7\suppep{1} & 45.8\suppep{1} & 46.7\suppep{4} & 62.4\suppstd{0.23} & 90.9\suppstd{0.15} & 95.7\suppstd{0.12} & 62.7\suppep{4} & 62.3\suppep{5} & 62.3\suppep{5} \\  %
    \rowcolor{oursbg} Ours + distillation & 47.1\suppstd{0.03} & 77.3\suppstd{0.30} & 85.7\suppstd{0.22} & 47.1\suppep{2} & 47.2\suppep{4} & 47.1\suppep{3} & 63.3\suppstd{0.06} & 91.5\suppstd{0.10} & 96.0\suppstd{0.20} & 63.3\suppep{4} & 63.2\suppep{5} & 63.3\suppep{4} \\  %
    \bottomrule
  \end{tabular}
\end{table*}

\begin{table*}[t]
  \caption{Per-seed results on LSMDC and DiDeMo (\cref{tab:standard}); format as in \cref{tab:supp_seeds_a}. Epoch chosen on the validation split. LSMDC: $5$ epochs. DiDeMo: $10$ epochs.}
  \label{tab:supp_seeds_e}
  \centering
  \footnotesize
  \setlength{\tabcolsep}{2.5pt}
  \begin{tabular}{lccc@{\hspace{6pt}}ccc@{\hspace{10pt}}ccc@{\hspace{6pt}}ccc}
    \toprule
    Method & \multicolumn{6}{c}{LSMDC} & \multicolumn{6}{c}{DiDeMo} \\
    \cmidrule(lr){2-7} \cmidrule(lr){8-13}
    & \multicolumn{3}{c}{mean $\pm$ std} & \multicolumn{3}{c}{R@1 per seed (epoch)} & \multicolumn{3}{c}{mean $\pm$ std} & \multicolumn{3}{c}{R@1 per seed (epoch)} \\
    \cmidrule(lr){2-4} \cmidrule(lr){5-7} \cmidrule(lr){8-10} \cmidrule(lr){11-13}
    & R@1 & R@5 & R@10 & 1 & 2 & 3 & R@1 & R@5 & R@10 & 1 & 2 & 3 \\
    \midrule
    CLIP4Clip-MeanP & 17.9\suppstd{0.75} & 38.4\suppstd{0.75} & 48.7\suppstd{0.53} & 18.7\suppep{3} & 17.2\suppep{2} & 17.9\suppep{3} & 39.4\suppstd{1.16} & 68.4\suppstd{0.20} & 78.5\suppstd{0.46} & 38.6\suppep{3} & 40.7\suppep{8} & 38.8\suppep{5} \\  %
    CLIP4Clip-seqTransf & 21.1\suppstd{0.47} & 40.8\suppstd{0.67} & 50.2\suppstd{0.96} & 20.7\suppep{2} & 21.6\suppep{2} & 20.9\suppep{2} & 40.4\suppstd{0.20} & 69.0\suppstd{0.75} & 79.3\suppstd{0.30} & 40.4\suppep{10} & 40.2\suppep{8} & 40.6\suppep{10} \\  %
    TeachCLIP w/o distill. & 21.6\suppstd{0.26} & 40.3\suppstd{0.96} & 48.9\suppstd{1.36} & 21.9\suppep{2} & 21.4\suppep{3} & 21.5\suppep{2} & 43.3\suppstd{0.61} & 70.8\suppstd{0.25} & 79.3\suppstd{0.31} & 42.6\suppep{10} & 43.6\suppep{8} & 43.7\suppep{9} \\  %
    X-CLIP & 21.9\suppstd{0.42} & 41.8\suppstd{0.62} & 51.0\suppstd{0.75} & 21.4\suppep{3} & 22.2\suppep{2} & 22.0\suppep{2} & 45.4\suppstd{0.65} & 73.7\suppstd{0.17} & 82.2\suppstd{0.45} & 45.4\suppep{9} & 44.7\suppep{10} & 46.0\suppep{9} \\  %
    X-Pool & 21.0\suppstd{0.21} & 40.5\suppstd{0.56} & 49.9\suppstd{0.56} & 20.8\suppep{3} & 21.2\suppep{3} & 20.9\suppep{3} & 44.6\suppstd{0.25} & 72.7\suppstd{0.42} & 82.2\suppstd{0.69} & 44.6\suppep{7} & 44.9\suppep{5} & 44.4\suppep{6} \\  %
    TS2-Net & 20.3\suppstd{0.21} & 38.8\suppstd{0.23} & 47.7\suppstd{0.26} & 20.1\suppep{2} & 20.5\suppep{2} & 20.4\suppep{2} & 39.1\suppstd{1.27} & 68.2\suppstd{0.25} & 79.3\suppstd{0.29} & 40.2\suppep{7} & 39.3\suppep{6} & 37.7\suppep{5} \\  %
    TeachCLIP & 21.1\suppstd{0.61} & 41.4\suppstd{0.85} & 50.5\suppstd{0.06} & 21.4\suppep{2} & 21.5\suppep{2} & 20.4\suppep{2} & 44.5\suppstd{0.51} & 71.0\suppstd{0.25} & 80.6\suppstd{0.12} & 44.9\suppep{8} & 44.6\suppep{8} & 43.9\suppep{8} \\  %
    \midrule
    \rowcolor{oursbg} Ours & 22.7\suppstd{0.45} & 42.9\suppstd{0.70} & 51.8\suppstd{0.55} & 22.7\suppep{2} & 22.2\suppep{2} & 23.1\suppep{2} & 43.3\suppstd{0.40} & 72.6\suppstd{1.21} & 81.7\suppstd{0.35} & 43.3\suppep{8} & 42.9\suppep{7} & 43.7\suppep{5} \\  %
    \rowcolor{oursbg} Ours + distillation & 22.5\suppstd{0.29} & 42.4\suppstd{0.38} & 52.2\suppstd{1.06} & 22.2\suppep{2} & 22.7\suppep{3} & 22.7\suppep{2} & 44.6\suppstd{0.47} & 72.4\suppstd{0.31} & 81.8\suppstd{0.51} & 44.4\suppep{5} & 44.2\suppep{9} & 45.1\suppep{9} \\  %
    \bottomrule
  \end{tabular}
\end{table*}

\section{Query-Time Cost and Gallery Size}
\label{sec:supp_cost}
This section details the cost numbers of \cref{fig:teaser}b and \cref{sec:experiments}: what we time, how we count operations, and how the gap grows with the gallery.

\paragraph{What we time.}
We time only the part of retrieval that depends on both the text and the video (\cref{fig:teaser}b and \cref{sec:experiments}).
The video side is computed offline and stored in an index.
The text queries are encoded beforehand, and text encoding is not timed.
For each method, the timed code goes from the stored index and the encoded queries to the top-10 list of each query.
It computes the full score matrix and then takes the top 10.
The scoring code follows the evaluation code of each method line by line.  %
All runs use one A40 GPU and fp16, with $T{=}12$ frames, $L{=}32$ words and $D{=}512$.  %
Texts are scored in tiles, as in our evaluation code: $100$ texts per tile for X-CLIP, $50$ for X-Pool, and all texts at once for the other methods.  %
Videos are scored in chunks to bound memory.
Each setting runs once as warm-up and then five times, and we report the median time divided by the number of queries.  %
We use $1{,}000$ queries ($100$ for the gallery of $1$M videos).  %

\paragraph{Gallery.}
The gallery does not hold real video features or copies of them.
It holds random unit vectors of the right shape, since the time does not depend on the values.  %
We measure $N = 1$k, $10$k, $100$k and $1$M videos.
\Cref{fig:teaser}b uses $N = 100$k.  %
The index size is the number of stored vectors per video $\times N \times D \times 2$ bytes.  %

\paragraph{Operations per text--video pair.}
We count the multiply-adds that depend on both the text and the video, divided by $D$, so that one $D$-dimensional dot product counts as $1$.  %
Work that depends on only the video or only the text is done once and is not counted.
\begin{itemize}
  \item \emph{One vector per clip} (CLIP4Clip-MeanP, CLIP4Clip-seqTransf, TeachCLIP w/o distillation, TeachCLIP and our method): one dot product, so the count is $1$.
  \item \emph{TS2-Net}: the score used at test time compares the text with all $12$ stored frame vectors and weights them with a softmax over frames that depends on the text.  %
  This costs $T{=}12$ dot products per pair, plus the softmax.
  TS2-Net therefore does not store one vector per clip.
  \item \emph{X-CLIP}: $1$ video--sentence, $L$ video--word, $T$ sentence--frame and $TL$ frame--word dot products, that is, $(T{+}1)(L{+}1) = 429$.
  The learned weight matrices on the similarity scores add $(L^2{+}T^2{+}TL^2{+}LT^2{+}T^2{+}L^2)/D \approx 37.6$, for a total of $466.6$.  %
  X-CLIP stores $13$ vectors per clip ($12$ frames and $1$ video vector).
  \item \emph{X-Pool}: X-Pool pools the frames again for each query.
  This takes $T$ attention scores and a weighted sum of $T$ values ($2T$), the output and the final linear projections ($2D$), and the final dot product ($1$): $2T{+}2D{+}1 = 1{,}049$.  %
  This count assumes that the key and value projections of the frames are computed offline.
  The official code instead recomputes them from the $12$ stored frame features for each tile of queries.
  We time both versions (\cref{tab:supp_cost}); \cref{fig:teaser}b uses the official one.  %
\end{itemize}
LayerNorm, softmax and scaling are not counted in these formulas.  %

\begin{table*}[t]
  \caption{Measured retrieval time per text query, from the stored index to the top-10 list (one A40, fp16, $T{=}12$, $L{=}32$, $D{=}512$; text encoding not timed). The gallery holds random unit vectors. Operations per pair: $D$-dimensional dot products that depend on both the text and the video. \Cref{fig:teaser}b uses the column $N{=}100$k and the official X-Pool.}
  \label{tab:supp_cost}
  \centering
  \small
  \setlength{\tabcolsep}{4pt}
  \begin{tabular}{lccccccc}
    \toprule
    & Vectors & Operations & \multicolumn{4}{c}{Time per query (ms), gallery size $N$} & Index (GB) \\
    \cmidrule(lr){4-7}
    Method & per clip & per pair & $1$k & $10$k & $100$k & $1$M & $N{=}100$k \\
    \midrule
    One vector per clip (CLIP4Clip-MeanP, Ours, \ldots) & 1 & 1 & 0.00014 & 0.00049 & 0.0037 & 0.049 & 0.10 \\
    TS2-Net & 12 & $\approx 12$ & 0.0010 & 0.0089 & 0.088 & 1.06 & 1.23 \\
    X-CLIP & 13 & 466.6 & 0.051 & 0.50 & 4.92 & 49.4 & 1.33 \\
    X-Pool (key/value offline) & 24 & 1{,}049 & 0.038 & 0.35 & 3.48 & 34.9 & 2.46 \\
    X-Pool (official) & 12 & $>1{,}049$ & 0.043 & 0.39 & 3.91 & 39.1 & 1.23 \\
    \bottomrule
  \end{tabular}
\end{table*}

\paragraph{How the gap grows with the gallery.}
For every method, the time grows about linearly with $N$ from $10$k videos on (\cref{tab:supp_cost}).
The gap in milliseconds thus grows with the gallery.
With one vector per clip, a query over $1$M videos takes $0.049$\,ms; with X-CLIP it takes $49.4$\,ms, and with the official X-Pool $39.1$\,ms.  %
At $N{=}100$k, X-CLIP is about $1{,}330\times$ slower than one vector per clip, the official X-Pool about $1{,}060\times$ and TS2-Net about $24\times$.  %
The ratio is smaller at $N{=}1$k (X-CLIP about $350\times$), where fixed overheads dominate the time of the single matrix product.  %
From $10$k videos on, it stays between about $800\times$ and $1{,}330\times$ for X-CLIP and the official X-Pool.  %
For X-Pool with keys and values computed offline, the measured ratio at $100$k ($943\times$) is close to the operation count ($1{,}049$); the official X-Pool needs more operations and measures $1{,}060\times$.  %
For X-CLIP, it is larger than the count ($1{,}330$ vs.\ $467$), since many of its steps (\eg, the softmax over frames and words) are limited by memory access rather than by arithmetic.  %
The index also grows with the number of stored vectors: $0.10$\,GB for $100$k clips with one vector each, and $1.2$ to $2.5$\,GB for the other methods.  %

\section{Number of Rounds}
\label{sec:supp_rounds}
This section supports the choice $R{=}2$ (\cref{sec:rounds}) and the round-drop study of \cref{sec:ablation}.

\begin{figure}[t]
  \centering
  \includegraphics[width=\linewidth]{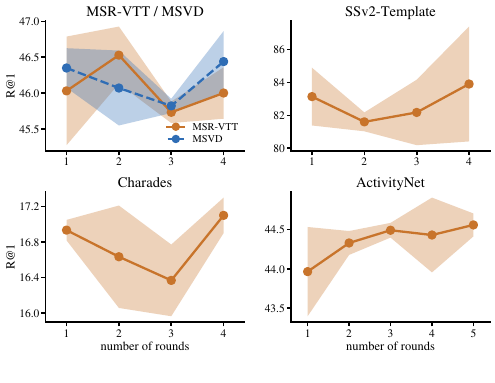}  %
  \caption{R@1 against the number of trained rounds (each round with its own parameters). Band: mean$\pm$std over three seeds. The first panel shows MSR-VTT (solid) and MSVD (dashed).}
  \label{fig:supp_rounds}
\end{figure}

\paragraph{R@1 against the number of rounds.}
\Cref{fig:supp_rounds} and \cref{tab:supp_rounds} give R@1 for models trained with $R=1$ to $4$ rounds (each round with its own parameters, \cref{sec:rounds}) on five benchmarks, and with $R=5$ on ActivityNet.
Every point is the mean over three seeds.
On SSv2-T, one, two and four rounds give $83.1$, $81.6$ and $83.9$.  %
These differences are within the variation across seeds (std $1.8$, $0.6$ and $3.5$).  %
On MSR-VTT, R@1 is highest with two rounds ($46.5$) and does not rise with three or four rounds ($45.7$ and $46.0$).  %
On MSVD, all values lie between $45.8$ and $46.4$.  %
On the two benchmarks with longer videos and several actions, more rounds help slightly.
Four rounds reach $17.1$ on Charades, against $16.6$ with two.  %
Five rounds reach $44.6$ on ActivityNet, against $44.3$ with two.
Four rounds on Charades and five on ActivityNet are chosen on the test set.  %

\paragraph{Why one $R$ for all benchmarks.}
The main text fixes $R=2$ on every benchmark and does not choose $R$ per benchmark (\cref{sec:rounds}).
SSv2, Charades, ActivityNet and MSR-VTT have no validation split in our protocol (\cref{sec:supp_setup}), so a per-benchmark choice of $R$ would be made on the test set.  %
Across $R$, the mean R@1 changes by at most $2.3$ on SSv2-T, $0.8$ on MSR-VTT, $0.7$ on Charades, $0.6$ on ActivityNet and $0.6$ on MSVD (\cref{tab:supp_rounds}). We therefore keep $R{=}2$ on every benchmark.  %

\begin{table}[t]
  \caption{R@1 (mean$\pm$std over three seeds) against the number of trained rounds $R$. The main tables use $R{=}2$. For ActivityNet, the std is computed at a higher precision than in \cref{tab:supp_seeds_c} ($0.15$ vs.\ $0.12$ for $R{=}2$).}  %
  \label{tab:supp_rounds}
  \centering
  \footnotesize
  \setlength{\tabcolsep}{3pt}
  \begin{tabular}{@{}lccccc@{}}
    \toprule
    Benchmark & $R{=}1$ & $R{=}2$ & $R{=}3$ & $R{=}4$ & $R{=}5$ \\
    \midrule
    SSv2-T      & 83.1\,{\scriptsize$\pm$1.8} & 81.6\,{\scriptsize$\pm$0.6} & 82.2\,{\scriptsize$\pm$2.0} & 83.9\,{\scriptsize$\pm$3.5} & -- \\  %
    Charades    & 16.9\,{\scriptsize$\pm$0.1} & 16.6\,{\scriptsize$\pm$0.6} & 16.4\,{\scriptsize$\pm$0.4} & 17.1\,{\scriptsize$\pm$0.2} & -- \\  %
    ActivityNet & 44.0\,{\scriptsize$\pm$0.6} & 44.3\,{\scriptsize$\pm$0.2} & 44.5\,{\scriptsize$\pm$0.1} & 44.4\,{\scriptsize$\pm$0.5} & 44.6\,{\scriptsize$\pm$0.1} \\  %
    MSR-VTT     & 46.0\,{\scriptsize$\pm$0.8} & 46.5\,{\scriptsize$\pm$0.4} & 45.7\,{\scriptsize$\pm$0.2} & 46.0\,{\scriptsize$\pm$0.4} & -- \\  %
    MSVD        & 46.4\,{\scriptsize$\pm$0.3} & 46.1\,{\scriptsize$\pm$0.5} & 45.8\,{\scriptsize$\pm$0.1} & 46.4\,{\scriptsize$\pm$0.4} & -- \\  %
    \bottomrule
  \end{tabular}
\end{table}

\paragraph{Dropping rounds at test time.}
\Cref{tab:supp_drop} gives the R@1 values behind \cref{fig:round_drop}.
We take the four-round model and keep only its first $k$ rounds at test time, without retraining.
With one round kept, SSv2-T loses $16.5$ R@1 on average; the loss is $10.9$ to $19.5$ for the three seeds.  %
\Cref{tab:supp_spread} gives, for each benchmark, the amount (\cref{eq:amount}) each round adds and how much its frame weights overlap with those of the previous round.  %
On every benchmark the later rounds read mostly the same frames as the round before (overlap $77$--$89\%$), so they do not move to other frames.
What differs is the amount: on SSv2-T each round adds about $0.33$, while on MSR-VTT the amount falls from $0.22$ to $0.12$ and on the other benchmarks it stays small ($0.06$--$0.19$).
These values are for the four-round model (three seeds, last epoch); the values in \cref{sec:why} ($0.64$ and $0.55$ on SSv2-T, $0.27$ and $0.18$ on MSR-VTT) are for one seed of the two-round model.  %

\begin{table}[t]
  \caption{Four-round model, test sets, last epoch: the amount each round adds (\cref{eq:amount}) and the overlap of its frame weights with those of the previous round (weights renormalized over the frames, without the null slot, to sum to one; the sum over frames of the smaller of the two weights, \ie, histogram intersection; \%). Mean over three seeds; std $\le 0.10$ for the amount and $\le 2.5$ for the overlap. MSVD is averaged over its test captions.}  %
  \label{tab:supp_spread}
  \centering\footnotesize\setlength{\tabcolsep}{4pt}
  \begin{tabular}{@{}lcccc|ccc@{}}
    \toprule
    & \multicolumn{4}{c|}{Amount, round} & \multicolumn{3}{c}{Overlap (\%), round} \\
    & 1 & 2 & 3 & 4 & 2 & 3 & 4 \\
    \midrule
    SSv2-T      & 0.34 & 0.32 & 0.33 & 0.34 & 77 & 79 & 78 \\
    Charades    & 0.16 & 0.19 & 0.17 & 0.14 & 84 & 82 & 82 \\
    ActivityNet & 0.09 & 0.07 & 0.07 & 0.09 & 87 & 86 & 85 \\
    MSR-VTT     & 0.22 & 0.18 & 0.17 & 0.12 & 89 & 89 & 88 \\
    MSVD        & 0.09 & 0.10 & 0.09 & 0.06 & 88 & 89 & 89 \\
    \bottomrule
  \end{tabular}
\end{table}

\begin{table}[t]
  \caption{R@1 of the four-round model when only its first rounds are kept at test time (no retraining; mean over three seeds). This is the data of \cref{fig:round_drop}.}
  \label{tab:supp_drop}
  \centering
  \footnotesize
  \setlength{\tabcolsep}{4pt}
  \begin{tabular}{@{}lcccc@{}}
    \toprule
    Rounds kept & 4 & 3 & 2 & 1 \\
    \midrule
    SSv2-T      & 83.9 & 79.3 & 75.5 & 67.4 \\  %
    Charades    & 17.1 & 16.9 & 16.3 & 15.2 \\  %
    ActivityNet & 44.4 & 44.1 & 43.8 & 43.4 \\  %
    MSR-VTT     & 46.0 & 45.9 & 46.1 & 45.4 \\  %
    MSVD        & 46.4 & 46.5 & 46.5 & 46.5 \\  %
    \bottomrule
  \end{tabular}
\end{table}

\section{Further Ablations}
\label{sec:supp_ablation}

\Cref{tab:supp_abl} lists variants that are not in \cref{tab:abl_parts}, and repeats the learned fixed query of \cref{tab:abl_parts} with its std and two more seeds.
Each row changes one part of our method and is compared with the reference row of its block.
Values are mean$\pm$std over three seeds unless stated.
MSR-VTT is trained for 5 epochs and SSv2-T for 20 epochs, as in the main tables.

\begin{table}[t]
  \caption{Further ablations (text-to-video R@1, mean$\pm$std over three seeds unless stated). Each row changes one part of the reference row above it. Shaded: reference rows. --: not run.}
  \label{tab:supp_abl}
  \centering
  \footnotesize
  \setlength{\tabcolsep}{4pt}
  \begin{tabular}{@{}lcc@{}}
    \toprule
    Variant & MSR-VTT & SSv2-T \\
    \midrule
    \rowcolor{black!6} Ours, $R{=}1$ (reference) & 46.0\,{\scriptsize$\pm$0.8} & 83.1\,{\scriptsize$\pm$1.8} \\  %
    Amount: scalar gate                 & 45.4\,{\scriptsize$\pm$0.4} & -- \\  %
    Amount: sigmoid gate                & 45.8\,{\scriptsize$\pm$1.0} & -- \\  %
    Values without LN and scale         & 45.2\,{\scriptsize$\pm$0.5} & -- \\  %
    Null bias initialized at $b{=}0$    & 45.6\,{\scriptsize$\pm$0.7} & -- \\  %
    Text query instead of $\mu$         & 46.3\,{\scriptsize$\pm$0.2} & 83.3\,{\scriptsize$\pm$0.0} \\  %
    Learned fixed query                 & 45.2\,{\scriptsize$\pm$0.1} & 82.8\,{\scriptsize$\pm$4.0} \\  %
    \quad five seeds                   & -- & 82.9\,{\scriptsize$\pm$2.9} \\  %
    \midrule
    \rowcolor{black!6} Ours, $R{=}2$ (reference) & 46.5\,{\scriptsize$\pm$0.4} & 81.6\,{\scriptsize$\pm$0.6} \\  %
    2 rounds, shared weights            & 45.8\,{\scriptsize$\pm$0.6} & 80.3\,{\scriptsize$\pm$1.4} \\  %
    1 round, 2 parallel heads           & 46.2\,{\scriptsize$\pm$0.4} & -- \\  %
    1 round, 4 parallel heads           & 45.9\,{\scriptsize$\pm$0.4} & -- \\  %
    Round 2: shifted candidates         & 46.2\,{\scriptsize$\pm$1.1} & 81.2\,{\scriptsize$\pm$2.4} \\  %
    Round 2: rebuilt candidates         & 46.4\,{\scriptsize$\pm$0.2} & 81.0\,{\scriptsize$\pm$3.0} \\  %
    \midrule
    \rowcolor{black!6} Ours, $R{=}3$ (reference) & 45.7\,{\scriptsize$\pm$0.2} & -- \\  %
    3 rounds, ReZero on round 3         & 45.5\,{\scriptsize$\pm$0.7} & -- \\  %
    \bottomrule
  \end{tabular}
\end{table}

\paragraph{How much to add (first block).}
In our method, the amount added to $\mu$ is set by the competition with the null slot (\cref{eq:amount}).
The two gate variants remove the null slot, take a softmax over the candidates only, and multiply the result by a gate.  %
The scalar gate is one learned scalar shared by all clips; it lowers R@1 by $0.7$.  %
The sigmoid gate computes a per-clip amount from $\mu$, $\sigma(w^\top\mathrm{LN}(\mu)+\beta)$, with learned $w$ and $\beta$; it is $0.2$ below the reference, which is within the seed std.  %
Removing the layer normalization and the $\lVert\mu\rVert/\sqrt{d}$ scale of the values ($u_t=W_v c_t$) lowers R@1 by $0.8$.  %
Without them, the value norms vary, so the null slot no longer sets the amount alone (\cref{sec:query}).
Initializing the null bias at $b=0$, instead of the value that adds about $10\%$ at the start with $12$ frames, lowers R@1 by $0.5$.  %

\paragraph{Which query (first block).}
Replacing the query $\mu$ by the text, as in X-Pool, gives about the same R@1 ($46.3$ vs.\ $46.0$ on MSR-VTT, $83.3$ vs.\ $83.1$ on SSv2-T).  %
The text query, however, needs the video vector to be recomputed for every query, while $\mu$ keeps it offline.
A learned fixed attention query, the same for every clip, lowers R@1 on MSR-VTT by $0.8$. The std is $0.1$ for the fixed query and $0.8$ for the reference, but every seed of the fixed query ($45.2$ to $45.3$) is below every seed of the reference ($45.5$ to $46.9$).  %
On SSv2-T, its seeds vary widely: $81.0$, $79.9$ and $87.4$ for the three seeds (\cref{tab:abl_parts} uses their mean, $82.8$).  %
Two more seeds give $83.9$ and $82.2$, so the five-seed mean is $82.9$, against $83.1$ for $\mu$.  %
On SSv2-T, the fixed query is thus close to $\mu$. On MSR-VTT, whose content varies from clip to clip, it helps that the attention query comes from the clip itself, as $\mu$ does.

\paragraph{How to use more parameters (second block).}
Sharing the weights of the two rounds, with only the null bias per round, lowers R@1 by $0.7$ on MSR-VTT and $1.3$ on SSv2-T (three seeds each; on SSv2-T within the seed std).  %
Parallel heads have the same parameters as extra rounds: each head has its own projections, layer normalizations, null key and bias, and is queried by $\mu$; their additions are summed.  %
On MSR-VTT, one round with two heads gives $46.2$, against $46.5$ for two rounds; with four heads, $45.9$ against $46.0$ for four rounds.  %
With the same parameters, parallel heads thus do not beat sequential rounds. In sequential rounds, each round is queried by the current vector.

\paragraph{Updating the candidates (second block).}
All rounds share the same candidates $c_t$ (\cref{sec:rounds}).
We test two ways to update them before round 2.
\emph{Shifted} subtracts what round 1 added, $c_t-\gamma\,(v^{(1)}-\mu)$, where $\gamma$ is the mean frame-embedding norm.
\emph{Rebuilt} runs the temporal transformer again. Its input is $x_t-\gamma\,v^{(1)}$, the deviation from the current vector, instead of $x_t-\bar{x}$.  %
Neither helps: both stay within the seed std of the reference on MSR-VTT and SSv2-T.  %

\paragraph{ReZero for a third round (third block).}
Three rounds are below two on MSR-VTT ($45.7$ vs.\ $46.5$).  %
We therefore scale the addition of round 3 by a learned scalar initialized at $0$, so that training starts from the two-round model.  %
This does not help: $45.5$ vs.\ $45.7$ for plain three rounds on MSR-VTT, and $16.3$ vs.\ $16.4$ on Charades (not in \cref{tab:supp_abl}).  %

\paragraph{Rows of \cref{tab:abl_parts}.}
The frame MLP adds a residual MLP to each frame output of the temporal transformer of TeachCLIP w/o distillation, before the AFA pooling: layer normalization, a $512{\to}768$ layer, GELU and a $768{\to}512$ layer whose weights start at zero.  %
It has about $0.79$M parameters, as many as one round of ours.  %
It changes R@1 little: $+0.6$ on SSv2-T and $-0.3$ on MSR-VTT, both within its seed std ($2.5$ and $0.4$).  %
The seed std of the other rows of \cref{tab:abl_parts} is, on SSv2-T and MSR-VTT: $1.8$ and $0.8$ for ours ($R{=}1$), $0.9$ and $0.3$ for candidates from $x_t$, $0.9$ and $0.9$ for no null slot, $1.1$ and $0.2$ for TeachCLIP w/o distillation, and $2.9$ and $0.4$ for input $x_t-\bar{x}$; the learned fixed query is in \cref{tab:supp_abl}.  %

\section{Distillation}
\label{sec:supp_distill}
This section gives the distillation setup of \cref{sec:experiments} and checks that our gain over TeachCLIP does not come from distillation.

\paragraph{Losses and weights.}
TeachCLIP and Ours + distillation use the same teachers and the same coarse-grained distillation loss of TeachCLIP~\cite{tian2024teachclip} (\cref{sec:experiments}).
This loss matches the student's video--text similarities to those of the teachers.
We keep the variant set in the TeachCLIP code (Pearson correlation).  %
The total loss is $0.2\,\mathcal{L}_\text{gt} + 1.0\,\mathcal{L}_\text{distill}$, where $\mathcal{L}_\text{gt}$ is the symmetric InfoNCE loss (text-to-video and video-to-text) on the ground-truth pairs (\cref{sec:supp_setup}).  %
TeachCLIP adds its fine-grained loss with weight $2.0$; Ours + distillation does not use it (weight $0$).  %
The fine-grained loss trains the frame weights of the AFA block of TeachCLIP to follow the teachers' text-to-frame similarities (cross-entropy); our method has no such frame weights.  %
All other settings, including the number of epochs, are those of the same model without distillation on each benchmark.  %
On MSR-VTT, the TeachCLIP row is the original implementation (10 epochs); elsewhere, it is its student and losses in our code.  %

\paragraph{Teachers per benchmark.}
\Cref{tab:supp_teachers} lists the teachers.
We choose the teachers as follows.  %
(1)~Where we obtained released TeachCLIP teacher checkpoints (MSR-VTT, MSVD, DiDeMo and VATEX), we use all of them, as TeachCLIP did.  %
(2)~On SSv2-T, SSv2-L, ActivityNet and LSMDC, we use only X-CLIP, the strongest baseline that we trained there.
The other two teachers of TeachCLIP, X-Pool and TS2-Net, are weaker than the students there, and a teacher weaker than its student is unlikely to help.  %
For example, on SSv2-T, X-Pool and TS2-Net reach $55.9$ and $69.1$ R@1, while the students reach $75.3$ (TeachCLIP w/o distillation) and $81.6$ (ours); on ActivityNet, both reach $40.1$, while the students reach $42.9$ and $44.3$.  %
(3)~On Charades and RTime, we use all three teachers of TeachCLIP (X-CLIP, TS2-Net and X-Pool), trained by us.  %
In every case, TeachCLIP and Ours + distillation share the same teachers, so their comparison does not depend on this choice.

\begin{table}[t]
  \caption{Teachers of TeachCLIP and Ours + distillation. Trained: a baseline that we trained (\cref{tab:specific,tab:standard}), from the first seed; on SSv2, each student uses the X-CLIP of its own seed.}
  \label{tab:supp_teachers}
  \centering
  \footnotesize
  \setlength{\tabcolsep}{3pt}
  \begin{tabular}{@{}lll@{}}
    \toprule
    Benchmark & Teachers & Source \\
    \midrule
    MSR-VTT & X-CLIP, TS2-Net, X-Pool & released by TeachCLIP \\
    MSVD & X-CLIP, TS2-Net, X-Pool & released by TeachCLIP \\
    DiDeMo & X-CLIP, TS2-Net & released (no X-Pool) \\
    VATEX & X-CLIP, TS2-Net & released (no X-Pool) \\
    ActivityNet & X-CLIP & trained, last epoch \\
    LSMDC & X-CLIP & trained, last epoch \\
    SSv2-T, SSv2-L & X-CLIP & trained, last epoch \\
    Charades & X-CLIP, TS2-Net, X-Pool & trained, last epoch \\
    RTime & X-CLIP, TS2-Net, X-Pool & trained, val.-selected epoch \\
    \bottomrule
  \end{tabular}
\end{table}  %

\paragraph{Does distillation explain our gain?}
On MSR-VTT, we add the same coarse-grained distillation, from the same three teachers, to TeachCLIP w/o distillation and to our model.  %
TeachCLIP w/o distillation goes from $44.6$ to $45.0$ R@1 ($+0.3$ from unrounded values), and our model from $46.5$ to $47.0$ ($+0.5$).  %
Both students thus gain about the same, and our model stays ahead by $2.1$ R@1.  %
Adding the fine-grained loss of TeachCLIP as well gives TeachCLIP in our code, which reaches $44.8$; the original TeachCLIP code gives $45.6$.  %
With the same teachers, Ours + distillation is ahead of TeachCLIP on five of the six standard benchmarks (\cref{tab:standard}).  %
On VATEX, the one exception, TeachCLIP reaches $63.7$ and Ours + distillation $63.3$, while ours without distillation reaches $62.4$.  %

\paragraph{One teacher on SSv2.}
On SSv2, X-CLIP is the only teacher.
On SSv2-T, the teacher, taken at its last epoch, reaches $79.1$ R@1 on average over seeds ($80.3$ at the best test epoch, \cref{tab:specific}), between the two students without distillation ($75.3$ for TeachCLIP w/o distillation and $81.6$ for ours).  %
Ours + distillation reaches $82.2$, and TeachCLIP $78.9$.  %
On SSv2-L, the teacher ($48.4$; $48.7$ at the best test epoch) is below our model ($51.5$), and distillation does not help our model ($50.7$); TeachCLIP reaches $48.4$.  %

\paragraph{Number of epochs.}
Longer training does not help distillation on MSR-VTT.
On MSR-VTT, Ours + distillation reaches $47.0$ R@1 with $5$ epochs, as in the main table, and $45.8$ with $20$ epochs (mean over three seeds).  %
The original TeachCLIP code also does not improve with $20$ instead of $10$ epochs on MSR-VTT ($45.4$ vs.\ $45.6$).  %

\section{Stronger Backbone: CLIP ViT-B/16}
\label{sec:supp_backbone}
This section expands the ViT-B/16 sentence of \cref{sec:experiments} to five benchmarks (X-CLIP on two of them), plus three more benchmarks without X-CLIP.  %
\begin{table*}[t]
  \caption{CLIP ViT-B/16 as the backbone (text-to-video R@1, mean over three seeds); ViT-B/32 numbers are in \cref{tab:specific,tab:standard}.
  Vec.: vectors stored per clip. X-CLIP was not run with ViT-B/16 on DiDeMo, LSMDC and VATEX. ms/query: measured retrieval time per query over $100$k videos (as in \cref{fig:teaser}b). Bold: best per column.}
  \label{tab:backbone}
  \centering
  \footnotesize
  \setlength{\tabcolsep}{4pt}
  \begin{tabular}{@{}lcrccccc@{}}
    \toprule
    Method & Vec. & ms/query & SSv2-T & MSR-VTT & DiDeMo & LSMDC & VATEX \\
    \midrule
    TeachCLIP w/o distill. & 1 & 0.004 & 81.0 & 46.9 & 46.4 & 24.2 & \textbf{66.9} \\  %
    X-CLIP & 13 & 4.9 & 82.0 & \textbf{49.5} & -- & -- & -- \\
    \rowcolor{oursbg} Ours & 1 & 0.004 & \textbf{83.7} & 47.5 & \textbf{47.6} & \textbf{26.2} & 66.7 \\  %
    \bottomrule
  \end{tabular}
\end{table*}

\Cref{tab:backbone} replaces CLIP ViT-B/32 with ViT-B/16 and keeps the rest of the recipe. X-CLIP with ViT-B/16 is our re-implementation, trained with the same code and recipe as the other ViT-B/16 runs.
The epoch of each run is chosen as in \cref{sec:supp_setup}: on validation for DiDeMo, LSMDC and VATEX, and the best test epoch on SSv2-T and MSR-VTT, which have no validation split.  %
On DiDeMo and LSMDC, our method stays ahead of TeachCLIP w/o distillation with ViT-B/16 ($+1.3$ and $+2.0$).
On VATEX, the two are level ($66.7$ vs.\ $66.9$).
We also trained both models with ViT-B/16 on Charades, MSVD and RTime (three seeds each; best test epoch on Charades, validation on MSVD and RTime; not in \cref{tab:backbone}).
On Charades, our method is slightly below TeachCLIP w/o distillation ($24.5$ vs.\ $24.9$; std $0.8$ and $0.7$).
On MSVD and RTime, the two are level ($49.4$ vs.\ $49.2$ and $37.5$ vs.\ $37.5$).  %
On SSv2-T, whose queries describe actions, our method still improves over TeachCLIP w/o distillation, the baseline of the same cost ($81.0 \to 83.7$), although by less than with ViT-B/32 ($+2.7$ vs.\ $+6.3$).
It also stays above X-CLIP ($83.7\pm1.3$ vs.\ $82.0\pm2.0$ over three seeds), which stores $13$ vectors per clip and is about $1{,}330$ times slower at query time ($N{=}100$k, \cref{sec:supp_cost}).  %
On MSR-VTT, whose queries mostly describe appearance, X-CLIP gains more from the stronger backbone and is ahead ($49.5$ vs.\ $47.5$). Measured against our re-implementation of X-CLIP with ViT-B/32 ($47.2$; \cref{tab:standard} uses the official code), X-CLIP gains $2.3$ and our method $0.9$.  %
As with ViT-B/32 (\cref{sec:supp_rounds}), more rounds do not close this gap.  %
Three and four rounds give $47.5$ and $47.7$ on MSR-VTT (three seeds), against $47.5$ with two.
This suggests that, with a stronger backbone, matching frames and words separately pays off for appearance queries, while one vector stays competitive on SSv2-T, whose queries ask about the change.

\section{Cross-Domain Transfer per Target}
\label{sec:supp_xdomain}
\Cref{tab:supp_xdomain_full} gives every cell behind the MSR-VTT average of \cref{tab:xdomain}.
Each model is trained on MSR-VTT and tested on the test split of each other benchmark, with that benchmark's own protocol. The one exception is X-CLIP on DiDeMo: its learned frame and word weights fix its input to the training setting ($12$ frames and $32$ words), so it is tested there with $12$ frames instead of $64$. The TeachCLIP and X-Pool models trained on MSR-VTT are runs in our code (the TeachCLIP student and losses; X-Pool with our batch size and epochs), with in-domain R@1 $44.8$ and $45.4$; they are not the original-code TeachCLIP and the official-recipe X-Pool of \cref{tab:standard}.  %
The checkpoint is chosen by the rule of the training benchmark, never on the target.

Among the models other than X-Pool, our method with or without distillation is best on DiDeMo, MSVD, VATEX and SSv2-Label.  %
On LSMDC, X-CLIP is best ($15.9$) and our method is sixth of the eight models other than X-Pool ($14.6$).  %
On SSv2-Template, every model stays between $4.8$ and $6.9$ R@1, so this target does not separate the methods; we therefore leave it out of the average in \cref{tab:xdomain}.  %
The last two columns repeat the SSv2-T and SSv2-L columns of \cref{tab:xdomain}.

\begin{table*}[t]
  \caption{\textbf{Cross-domain retrieval per target} (text-to-video R@1, mean over three seeds).
  Avg.\ of 5: mean over DiDeMo, MSVD, VATEX, LSMDC and SSv2-Label, as in \cref{tab:xdomain}; SSv2-Template is left out of it.
  Gray: X-Pool, which pools the frames with the text query; bold: best among the other rows.}
  \label{tab:supp_xdomain_full}
  \centering
  \footnotesize
  \setlength{\tabcolsep}{4.5pt}
  \begin{tabular}{@{}lcccccc|c|cc@{}}
    \toprule
    Train & \multicolumn{7}{c|}{MSR-VTT} & SSv2-T & SSv2-L \\
    Test & DiDeMo & MSVD & VATEX & LSMDC & SSv2-L & SSv2-T & avg.\ of 5 & MSR-VTT & MSR-VTT \\
    \midrule
    CLIP4Clip-MeanP        & 28.5 & 38.8 & 47.4 & 8.9  & 14.6 & 5.0 & 27.7 & 15.9 & 25.8 \\
    CLIP4Clip-seqTransf    & 31.5 & 43.3 & 49.3 & 15.0 & 14.2 & 6.7 & 30.7 & 15.3 & 24.9 \\
    TeachCLIP w/o distill. & 30.7 & 43.3 & 49.8 & 14.7 & 14.5 & 5.9 & 30.6 & 12.2 & 25.2 \\
    X-CLIP                 & 32.4 & 44.3 & 51.0 & \textbf{15.9} & 15.4 & \textbf{6.9} & 31.8 & 12.4 & 24.7 \\
    {\color{black!30}X-Pool} & {\color{black!30}38.9} & {\color{black!30}46.5} & {\color{black!30}54.9} & {\color{black!30}16.4} & {\color{black!30}17.1} & {\color{black!30}6.1} & {\color{black!30}34.8} & {\color{black!30}20.3} & {\color{black!30}29.7} \\
    TS2-Net                & 30.0 & 41.6 & 49.0 & 11.5 & 13.6 & 6.1 & 29.1 & 13.0 & \textbf{26.7} \\
    TeachCLIP              & 31.0 & 44.2 & 50.8 & 14.9 & 15.0 & 5.4 & 31.2 & 15.6 & 25.7 \\
    \midrule
    \rowcolor{oursbg} Ours                & 33.3 & 44.6 & 51.5 & 14.6 & 15.5 & 4.8 & 31.9 & \textbf{17.5} & 25.6 \\
    \rowcolor{oursbg} Ours + distillation & \textbf{33.4} & \textbf{45.1} & \textbf{52.3} & 15.0 & \textbf{16.4} & 5.6 & \textbf{32.4} & 16.2 & 26.1 \\
    \bottomrule
  \end{tabular}
\end{table*}

\section{Frame Similarity and Frame Order}
\label{sec:supp_similarity}
\Cref{sec:frames} compares how similar the frames of one clip are with how similar the words of one query are.
We use pretrained CLIP ViT-B/32 with no trained head.
For a clip, we average the cosine similarity over all pairs of its frame embeddings ($12$ on MSR-VTT and SSv2).
For a query, we average it over all pairs of its word features from the last layer of the CLIP text encoder (start, end and padding tokens excluded).
Before the cosine, we subtract the mean of each modality over the dataset.
After this centering, two elements from different items score about zero ($|\cos|<0.005$ for frames and for words on both benchmarks).  %

\Cref{tab:supp_similarity} gives the values with and without centering.
Without centering, all values are high, because the CLIP space is anisotropic: even frames of different clips score $0.47$ on MSR-VTT and $0.66$ on SSv2.  %
After centering, frames of the same clip stay highly similar, while words of the same query do not.
Frames are about $3.5\times$ more similar to each other than words are on both benchmarks, and $2.6$ to $4.6\times$ on the seven other benchmarks (\cref{tab:supp_similarity}).  %

\begin{table}[t]
  \caption{\textbf{Mean cosine similarity within one item} in pretrained CLIP ViT-B/32: between the frames of a clip ($12$ on MSR-VTT and SSv2) and between the words of its query. Centered: after subtracting the dataset mean of each modality. MSR-VTT: 1k-A test. SSv2: the 2,088 test videos with their SSv2-Label captions. Other benchmarks: up to 1,000 test videos each, with the frames our models read (from 5 on LSMDC to 55 on ActivityNet on average) and one query per video. Ratio: centered frames divided by centered words, from the unrounded values.}
  \label{tab:supp_similarity}
  \centering
  \footnotesize
  \setlength{\tabcolsep}{4pt}
  \begin{tabular}{@{}lcc|cc|c@{}}
    \toprule
    & \multicolumn{2}{c|}{without centering} & \multicolumn{2}{c|}{centered} & \\
    Benchmark & frames & words & frames & words & ratio \\
    \midrule
    MSR-VTT & 0.82 & 0.53 & 0.64 & 0.18 & 3.50 \\  %
    SSv2    & 0.90 & 0.55 & 0.71 & 0.20 & 3.53 \\  %
    \midrule  %
    DiDeMo      & 0.87 & 0.51 & 0.72 & 0.26 & 2.77 \\
    MSVD        & 0.88 & 0.50 & 0.73 & 0.16 & 4.61 \\
    VATEX       & 0.88 & 0.50 & 0.72 & 0.20 & 3.60 \\
    LSMDC       & 0.88 & 0.55 & 0.62 & 0.19 & 3.25 \\
    ActivityNet & 0.81 & 0.49 & 0.61 & 0.24 & 2.55 \\
    Charades    & 0.88 & 0.51 & 0.61 & 0.16 & 3.91 \\
    RTime       & 0.89 & 0.50 & 0.74 & 0.23 & 3.21 \\
    \bottomrule
  \end{tabular}
\end{table}

\begin{figure}[t]
  \centering
  \includegraphics[width=\linewidth]{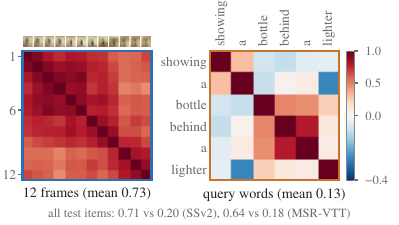}  %
  \caption{\textbf{Frames are similar, words are not.} Centered cosine similarity between the 12 frames of an SSv2 test clip (left) and between the words of its query, \emph{showing a bottle behind a lighter} (right), in pretrained CLIP ViT-B/32. Zero means as similar as two unrelated items. The clip is typical: its frame similarity is the median of the test set. The line at the bottom gives the means over all test items (\cref{tab:supp_similarity}).}
  \label{fig:supp_framesim}
\end{figure}

\Cref{fig:supp_framesim} shows one SSv2 test clip and its query.
We chose the clip by rule: among the clips whose caption has 5 to 9 words, it is the one whose frame similarity is closest to the median of the test set ($0.73$).  %
Its frames score $0.73$ and its words $0.13$.  %
Only a few word pairs, such as \emph{behind} and \emph{a}, are highly similar.  %

\paragraph{Deviation share and $\rho$.}
This paragraph gives the details of two numbers in \cref{sec:seqtransf}.
For the share of the deviations, we use the features and test clips of \cref{tab:supp_similarity}, that is, pretrained CLIP ViT-B/32, not the fine-tuned encoder.
We subtract the dataset mean and normalize each frame embedding to unit length; here, $x_t$ and its clip mean $\bar{x}$ denote these normalized embeddings.
The share of the deviations is then $\frac{1}{T}\sum_t\lVert x_t-\bar{x}\rVert^2=1-\lVert\bar{x}\rVert^2$.
Averaged over clips, it is $0.33$ on MSR-VTT and $0.27$ on SSv2.  %
We compute $\rho$ (\cref{eq:rho}) for each test clip with TeachCLIP w/o distillation, using the checkpoints of the main tables.  %
Averaged over clips and three seeds, it is $0.98$ on MSR-VTT, $0.96$ on SSv2-L and $0.94$ on SSv2-T, with a standard deviation over seeds of at most $0.007$.  %

\paragraph{Shuffled and reversed frames.}
\label{sec:supp_order}
This completes the order test of \cref{sec:direction}.
We re-evaluate the checkpoints of the main tables (three seeds) with the frames of each test clip intact, shuffled or reversed; for shuffled clips, we average over five random frame orders.  %
The four temporal models are CLIP4Clip-seqTransf, TeachCLIP w/o distillation, TeachCLIP w/o distillation with input $x_t-\bar{x}$ (\cref{tab:abl_parts}) and ours; \cref{tab:supp_order} gives all values.  %
On SSv2-T, reversing the frames lowers R@1 by $39.3$ to $43.5$ for these models.  %
Our gain over TeachCLIP w/o distillation is $6.3$ on intact clips, $5.5$ on shuffled clips and $2.1$ on reversed clips.  %
On SSv2-L, it is $3.5$ on intact clips, $1.1$ on shuffled clips and $-2.4$ on reversed clips, so on SSv2-L most of the gain depends on the frame order.  %
Thus, part of the gain depends on the direction of time, not only on a better summary of the frames.
These runs re-evaluate the stored checkpoints, so the intact values differ from \cref{tab:specific} by at most $0.2$ R@1.  %

\begin{table}[t]
  \caption{\textbf{Frame order} (text-to-video R@1, mean over three seeds). The test frames are intact, shuffled (shuf.; mean over five random orders) or reversed (rev.); checkpoints of the main tables, re-evaluated. Gain: ours minus TeachCLIP w/o distillation, from unrounded means. --: not run.}
  \label{tab:supp_order}
  \centering
  \footnotesize
  \setlength{\tabcolsep}{3pt}
  \begin{tabular}{@{}lccc|ccc@{}}
    \toprule
    & \multicolumn{3}{c|}{SSv2-T} & \multicolumn{3}{c}{SSv2-L} \\
    Model & intact & shuf. & rev. & intact & shuf. & rev. \\
    \midrule
    CLIP4Clip-seqTransf         & 78.7 & 59.3 & 36.0 & 48.1 & 44.4 & 39.8 \\  %
    TeachCLIP w/o distill.      & 75.5 & 58.2 & 36.2 & 48.0 & 43.9 & 39.4 \\  %
    \quad + input $x_t-\bar{x}$ & 82.2 & 64.4 & 39.1 & --   & --   & --   \\  %
    \rowcolor{oursbg} Ours      & 81.8 & 63.7 & 38.3 & 51.5 & 44.9 & 37.0 \\  %
    \midrule
    Gain of ours                & 6.3  & 5.5  & 2.1  & 3.5  & 1.1  & $-2.4$ \\
    \bottomrule
  \end{tabular}
\end{table}

\section{Qualitative Examples}
\label{sec:supp_examples}
\begin{figure*}[t]
  \centering
  \includegraphics[width=\linewidth]{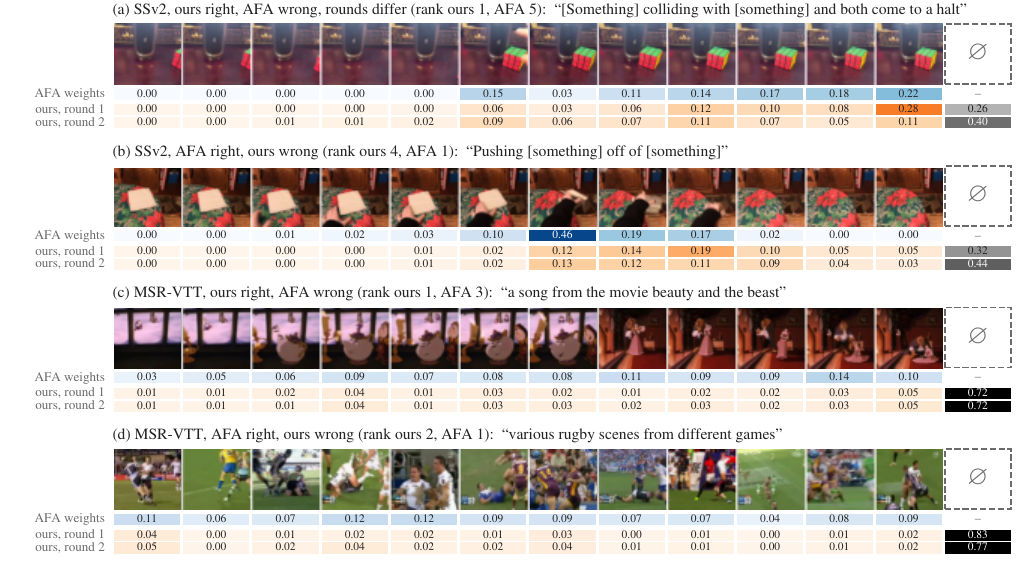}  %
  \caption{\textbf{More examples of where AFA and our method put their weight} (SSv2-T and MSR-VTT, first seed), in the style of \cref{fig:attn}.
  Blue: the AFA weights of TeachCLIP w/o distillation over the 12 frames.
  Orange: the weights of our two rounds over the frames and the null slot ($\varnothing$).
  (a)~SSv2-T, won by our method. \emph{[Something] colliding with [something] and both come to a halt}: our method ranks the video first, AFA fifth.
  Round one puts its largest weight on the last frame, where the objects have stopped; round two adds the frames before it more evenly, and gives more weight to the null slot ($0.40$ against $0.26$).  %
  (b)~SSv2-T, lost by our method. \emph{Pushing [something] off of [something]}: AFA ranks the video first, our method fourth.
  AFA keeps the moment of the push ($0.46$ on one frame), while our method spreads its weight over that moment and the frames after it.
  (c)~MSR-VTT, won by our method. \emph{a song from the movie beauty and the beast}: our method ranks the video first, AFA third.
  AFA spreads its weight over all frames ($0.03$ to $0.14$); both our rounds put $0.72$ on the null slot and at most $0.05$ on a frame.
  (d)~MSR-VTT, lost by our method. \emph{various rugby scenes from different games}: AFA ranks the video first, our method second.
  AFA again spreads its weight ($0.04$ to $0.12$); our rounds put $0.83$ and $0.77$ on the null slot.
  The rules that choose the queries are given in \cref{sec:supp_examples}.}
  \label{fig:supp_attn}
\end{figure*}

This section gives the selection rule behind \cref{fig:attn} and four more examples, including two losses.
\paragraph{How we choose the examples.}
We choose the examples by the rules below.  %
We compare our method with the AFA block of TeachCLIP w/o distillation, both with the first seed at the checkpoints of the main tables.
For each query, we take the rank of its best-ranked positive video under each model.
Let $W$ be the queries that our method ranks first and AFA does not, and $L$ the queries that AFA ranks first and our method does not.
On SSv2-T, $W$ has $24$ of the $174$ queries and $L$ has $12$.  %
On MSR-VTT, $W$ has $71$ of the $1{,}000$ queries and $L$ has $57$.  %
In each panel we show the positive video ranked best by the model that wins.
The examples are chosen by these rules:
\begin{itemize}
  \item \cref{fig:attn}a: from $W$ on SSv2-T, the query whose AFA weights have the median entropy.
  \item \cref{fig:attn}b: from $W$ on MSR-VTT, restricted to the $27$ queries that CLIP4Clip-MeanP (the frame mean, without temporal modeling) also ranks first, so that appearance alone answers them; among the eight whose null-slot weight, averaged over the two rounds, is closest to the median, we pick by eye one whose frames show a single scene.  %
  \item \cref{fig:supp_attn}a: from $W$ on SSv2-T, among the videos where both rounds add at least their median amount, the one where the frame weights of the two rounds overlap least (overlap: the sum over frames of the smaller of the two weights, as in \cref{sec:ablation}).
  \item \cref{fig:supp_attn}b: from $L$ on SSv2-T, the query whose AFA weights have the median entropy.
  \item \cref{fig:supp_attn}c and d: from $W$ (c) and from $L$ (d) on MSR-VTT, the query whose AFA weights have the median entropy.
\end{itemize}
Videos already shown in another panel are excluded.

\paragraph{More examples.}
\Cref{fig:supp_attn}a shows that two rounds can take different parts of an event.
Round one puts its largest weight on the last frame, where both objects have stopped ($0.28$).  %
Round two spreads its weight more evenly over frames 6 to 12, where the cube comes in ($0.05$ to $0.11$ each).  %
This example is chosen for the least overlap, so it is not typical: on average, the rounds of the four-round model overlap by about $80\%$ on SSv2-T (\cref{sec:ablation} and \cref{tab:supp_spread}).  %
\Cref{fig:supp_attn}b is a query that our method loses (rank $4$ against rank $1$ for AFA).  %
AFA puts $0.46$ of its weight on the single frame where the hand pushes the object off.  %
Our method spreads its weight over this frame and the frames after it ($0.09$ to $0.19$ each over frames 7 to 10).  %
When one moment decides the query, putting the weight on that moment can work better than adding the change to $\mu$.
On MSR-VTT (\cref{fig:supp_attn}c and d), AFA spreads its weight over all frames, and our method puts most of its weight on the null slot, whether it wins or loses.  %

\end{document}